\documentclass[10pt,twocolumn,letterpaper]{article}

\usepackage[pagenumbers]{cvpr} 

\usepackage[dvipsnames]{xcolor}

\definecolor{cvprblue}{rgb}{0.21,0.49,0.74}
\definecolor{green}{rgb}{0.18,0.55,0.34}
\usepackage[pagebackref,breaklinks,colorlinks,citecolor=green]{hyperref}

\usepackage{pgfplots} 

\usepackage{tikz}
\usepackage{xcolor}
\definecolor{blue}{RGB}{12, 58, 173}
\definecolor{red}{RGB}{207,78,56}
\definecolor{gray}{RGB}{146,146,161}
\definecolor{green4}{RGB}{46, 139, 87}
\definecolor{green2}{RGB}{102,153,39}
\colorlet{green2}{green2!80}
\usepackage{multirow} 
\usepackage{colortbl}
\definecolor{lightgray}{gray}{0.9}
\usepackage{hhline} 
\usepackage{pdflscape}

\usepackage{amsmath}
\usepackage{graphicx}
\usepackage{pifont}
\usepackage{dsfont}

\usepackage{amsthm}
\newtheorem{proposition}{Proposition}
\definecolor{pink}{RGB}{46, 139, 87}
\definecolor{red2}{RGB}{149,9,30}
\definecolor{barriercolor}{RGB}{255, 120, 50}
\definecolor{bicyclecolor}{RGB}{255, 192, 203}
\definecolor{buscolor}{RGB}{255, 255, 0}
\definecolor{carcolor}{RGB}{0, 150, 245}
\definecolor{constructcolor}{RGB}{0, 255, 255}
\definecolor{motorcolor}{RGB}{200, 180, 0}
\definecolor{pedestriancolor}{RGB}{255, 0, 0}
\definecolor{trafficcolor}{RGB}{255, 240, 150}
\definecolor{trailercolor}{RGB}{135, 60, 0}
\definecolor{truckcolor}{RGB}{160, 32, 240}
\definecolor{driveablecolor}{RGB}{255, 0, 255}
\definecolor{otherflatcolor}{RGB}{139, 137, 137}
\definecolor{sidewalkcolor}{RGB}{75, 0, 75}
\definecolor{terraincolor}{RGB}{150, 240, 80}
\definecolor{manmadecolor}{RGB}{213, 213, 213}
\definecolor{vegetationcolor}{RGB}{0, 175, 0}
\definecolor{otherscolor}{RGB}{0, 0, 0}

\providecommand{\ie}{\textit{i.e.}}
\providecommand{\eg}{\textit{e.g.}}
\providecommand{\vs}{\textit{v.s.}}
\providecommand{\wrt}{\textit{w.r.t.}}
\providecommand{\etc}{\textit{etc.}}

\usepackage{algorithm}
\usepackage{algpseudocode}
\usepackage[switch]{lineno}
\newcommand{\comment}[1]{\textcolor{gray}{\#~\mathit{#1}}}

\usepackage{setspace}

\newcommand{\crefcolor}[2]{%
  \begingroup
  \hypersetup{linkcolor=#1}%
  \cref{#2}%
  \endgroup
}
\newcommand{\refcolor}[2]{%
  \begingroup
  \hypersetup{linkcolor=#1}%
  \ref{#2}%
  \endgroup
}

\usepackage{tikz}

\def\paperID{****} 
\def\confName{CVPR}
\def\confYear{2025}

\title{Rethinking Temporal Fusion with a Unified Gradient Descent View for \\ 3D Semantic Occupancy Prediction} 

\author{
Dubing Chen$^{1}$,
Huan Zheng$^{1}$,
Jin Fang$^{1}$,
Xingping Dong$^2$, \\
Xianfei Li$^3$, 
Wenlong Liao$^3$, 
Tao He$^3$, 
Pai Peng$^3$,
Jianbing Shen$^{1}$\thanks{Corresponding author: \textit{Jianbing Shen}.
This work was supported in part by the Science and Technology Development Fund of Macau SAR (FDCT) under grants 0102/2023/RIA2 and 0154/2022/A3 and 001/2024/SKL, the Jiangyin Hi-tech Industrial Development Zone under the Taihu Innovation Scheme (EF2025-00003-SKL-IOTSC), the University of Macau SRG2022-00023-IOTSC grant, and the State Key Laboratory of Internet of Things for Smart City (University of Macau) (Ref. No.: SKL-IoTSC(UM)-2024-2026/ORP/GA04/2023).
}\\
$^1$SKL-IOTSC, CIS, University of Macau\\
$^2$Wuhan University, \hspace{4mm}    $^3$COWAROBOT Co. Ltd. \\
Project Page: \url{https://cdb342.github.io/GDFusion}
}

\begin{document}
\maketitle
\begin{abstract}
We present GDFusion, a temporal fusion method for vision-based 3D semantic occupancy prediction (VisionOcc). GDFusion opens up the underexplored aspects of temporal fusion within the VisionOcc framework, focusing on both temporal cues and fusion strategies. It systematically examines the entire VisionOcc pipeline, identifying three fundamental yet previously overlooked temporal cues: scene-level consistency, motion calibration, and geometric complementation. These cues capture diverse facets of temporal evolution and make distinct contributions across various modules in the VisionOcc framework.
To effectively fuse temporal signals across heterogeneous representations, we propose a novel fusion strategy by reinterpreting the formulation of vanilla RNNs. This reinterpretation leverages gradient descent on features to unify the integration of diverse temporal information, seamlessly embedding the proposed temporal cues into the network. Extensive experiments on nuScenes demonstrate that GDFusion significantly outperforms established baselines. Notably, on Occ3D benchmark, it achieves 1.4\%-4.8\% mIoU improvements and reduces memory consumption by 27\%-72\%.

\end{abstract}    
\vspace{1.5ex}
\section{Introduction}
\vspace{0.5ex}
\begin{figure*}[t]
    \centering
    \setlength{\abovecaptionskip}{0pt}
    \includegraphics[width=1\linewidth]{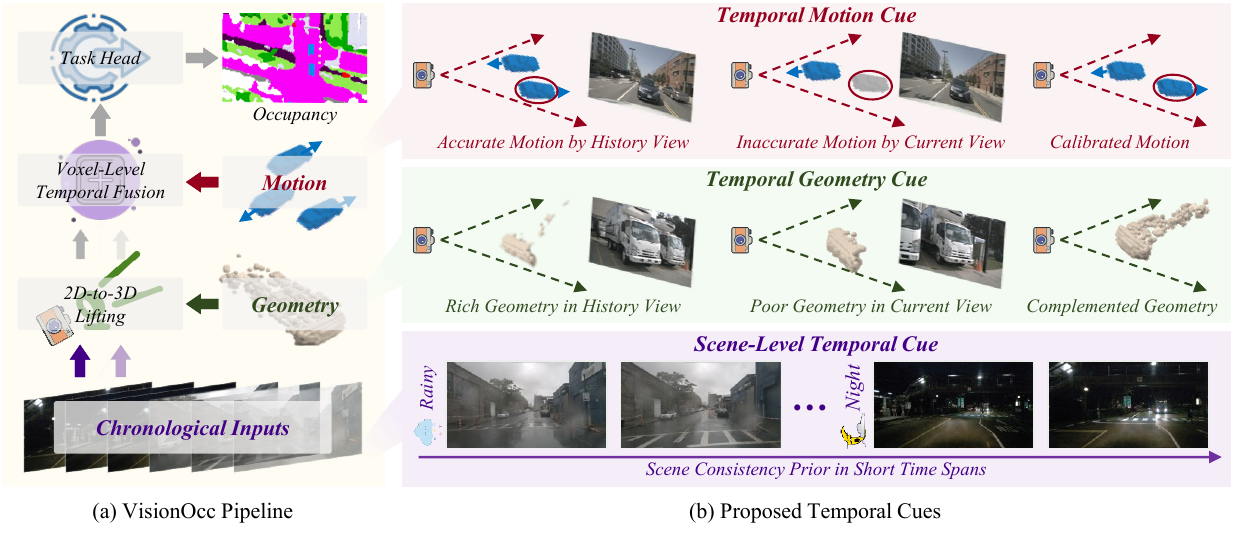}
   \vspace{-5mm}
    \caption{\textbf{Motivation behind the proposed temporal fusion.} \textbf{(a):} VisionOcc pipeline. \textbf{(b):} Proposed temporal cues, showing historical motion and geometric data enhancing current viewpoints, with scene consistency priors from historical information.}
    \label{fig:intro}
    \vspace{-1mm}
\end{figure*}

Dynamic environment perception and understanding play crucial roles in applications like autonomous driving and robotics \cite{elfes1989using,hu2023planning,jiang2023vad}. A key aspect of this understanding is 3D semantic occupancy prediction \cite{tong2023scene,tian2024occ3d,wang2023openoccupancy,huang2023tri}, which provides a detailed representation of both the spatial layout and semantic properties of the environment.

Given the continuous nature of dynamic perception tasks, temporal information is increasingly recognized as a key enhancer of performance \cite{park2022time,huang2022bevdet4d,wang2023exploring,han2024exploring}. By exploiting multi-frame data, these systems attain improved consistency and robustness in predictions. However, in vision-based 3D semantic occupancy prediction (VisionOcc), the potential of temporal fusion mechanisms remains largely underexplored. In this work, we conduct a systematic investigation into the temporal dynamics of VisionOcc, with a focus on analyzing comprehensive temporal cues and their unified integration within an RNN-style framework.

In general VisionOcc frameworks (\cref{fig:intro}\textcolor{red}{a}), image features are first transformed into volumetric features via a 2D-to-3D lifting module and decoded into occupancy predictions using task heads. For temporal integration, existing methods \cite{li2023fbocc,ma2023cotr,liu2024surroundsdf,chen2024alocc} primarily focus on voxel-level feature fusion. This intuitively enriches each voxel with additional temporal context to resolve ambiguities. Yet beyond this, we identify three crucial but unexplored temporal cues, as shown in \cref{fig:intro}\textcolor{red}{b}, each capturing distinct temporal aspects in VisionOcc: \textbf{i) Scene-level temporal cue:} Autonomous driving systems must adapt to various road conditions, weather patterns, and lighting situations, requiring robust domain generalization capabilities. Since scene conditions remain stable over short time spans, historical data provides valuable scene-specific cues (such as consistent environmental priors) that have been overlooked in prior research. \textbf{ii) Temporal motion cue:} In voxel-level temporal fusion, historical volumetric features must be aligned with the current frame’s coordinate system, with dynamic objects requiring additional motion compensation. While mispredictions of motion can occur in the current frame, the potential of leveraging historical motion information to correct these errors remains untapped. \textbf{iii) Temporal geometry cue:} Advanced VisionOcc methods utilize geometric priors (\eg, depth) \cite{huang2022bevdet4d,li2023fbocc,ma2023cotr,chen2024alocc} for 2D-to-3D lifting. Historical geometric information could effectively compensate for the predicted geometry of the current frame. 

These cues drive comprehensive temporal fusion in VisionOcc by fulfilling diverse roles in its pipeline. Scene-level fusion \textbf{captures information distinct from voxel-level fusion}, reflecting broader contextual cues. Temporal motion fusion \textbf{refines the accuracy of voxel-level fusion} by correcting alignment errors. Unlike other fusion strategies that target post-lifting 3D features, temporal geometry fusion improves current-frame 2D-to-3D lifting with historical geometry. Each type of temporal information has a distinct representation. As shown in \cref{fig:framework}, voxel-level information is encoded in 3D volume feature maps; scene-level information is represented as network parameters, inspired by test-time adaptation \cite{pmlr-v119-sun20b,ijcai2024p70} and large language models \cite{sun2024learning}; motion is captured via 3D flow maps; and geometry is encoded as probabilistic point clouds (detailed in \cref{sec:temp}). 

To integrate temporal information from heterogeneous representations, we propose a unified fusion framework, GDFusion. We seek to model comprehensive temporal fusion through efficient RNN processes, tailored to resource-constrained tasks like autonomous driving. However, vanilla RNNs struggle to embed temporal priors (scene-level information) into \textbf{network parameters}. They also fail to explicitly capture \textbf{motion dynamics} (motion information) and \textbf{probabilistic geometry} (geometry information) during fusion. To address this, we deconstruct the standard RNN by reinterpreting it as an optimization process. Specifically, it minimizes the MSE loss between current and historical features via gradient descent on the feature space. Based on this, we design specific loss functions to quantify discrepancies between each temporal representation and its current-frame counterpart. The gradients of these losses serve as temporal residuals, which are added to the corresponding representation to produce fused features. Through this mechanism, we effectively model all temporal fusions within an RNN-style structure, enabling efficient and adaptive integration of diverse temporal information.

GDFusion acts as a plug-and-play enhancement for VisionOcc methods. Extensive experiments across multiple baselines validate its robustness and broad applicability. Moreover, GDFusion showcases marked memory efficiency gains over traditional temporal fusion techniques like SOLOFusion \cite{park2022time}. The main contributions of this work are fourfold: \textbf{i)} We conduct an in-depth investigation of the temporal dynamics within the VisionOcc pipeline, uncovering three previously unexplored cues: scene-level, motion, and geometry. These cues respectively enable the incorporation of scene consistency priors, facilitate voxel-level temporal fusion, and enhance the 2D-to-3D feature transformation. \textbf{ii)} We deconstruct the vanilla RNN, reinterpreting it as gradient descent within the feature space. This allows the RNN to operate on a wide array of diverse representation forms. \textbf{iii)} Through this reinterpretation, we propose a unified RNN-style temporal fusion framework that efficiently integrates each type of temporal representation into the VisionOcc pipeline. \textbf{iv)} We evaluated GDFusion on three occupancy benchmarks. On Occ3D, it achieves 1.4\%-4.8\% mIoU gains over baselines with temporal fusion and reduces memory use by 27\%-72\%. On SurroundOcc and OpenOccupancy, it yields 6.3\%-14.0\% mIoU improvements over non-temporal baselines with negligible inference overhead.

\section{Related Work}
\vspace{-0.5ex}
\subsection{Vision-Based 3D Semantic Occupancy}
\vspace{-0.5ex}
Vision-based 3D semantic occupancy prediction estimates voxelized semantic states of 3D environments from visual inputs~\cite{zhang2024vision,tian2024occ3d,wang2023openoccupancy,wei2023surroundocc,huang2023tri}. MonoScene~\cite{cao2022monoscene} pioneers this by lifting features and using 3D context priors for spatial consistency. SurroundOcc~\cite{wei2023surroundocc}, Occ3D \cite{tian2024occ3d}, and OpenOccupancy \cite{wang2023openoccupancy} build dense occupancy annotations for autonomous driving by stacking multiple frames of sparse LiDAR point clouds. Studies like TPVFormer~\cite{huang2023tri}, FlashOcc~\cite{yu2023flashocc}, COTR~\cite{ma2023cotr}, SparseOcc \cite{liu2023fully}, and OSP~\cite{shi2024occupancy} explore efficient occupancy representations or acceleration techniques. TPVFormer introduces tri-perspective view representation, FlashOcc proposes height-to-channel compression, 
COTR proposes encoding on downsampled features, SparseOcc focuses on occupancy pruning, and OSP develops resolution-independent occupancy prediction. UniOcc \cite{pan2023uniocc}, OccNeRF~\cite{zhang2023occnerf}, GaussianOcc~\cite{gan2024gaussianocc}, \etc leverage techniques such as NeRF and 3D Gaussian Splatting to perform 3D semantic occupancy prediction with 2D annotations. OCCWorld \cite{zheng2025occworld}, OCCSora \cite{wang2024occsora}, ViewFormer \cite{li2024viewformer}, and ALOcc \cite{chen2024alocc} focus on dynamic occupancy prediction tasks, varying from future frame occupancy to occupancy flow prediction. Additionally, works like OCCWorld \cite{zheng2025occworld} and OccLLaMA \cite{wei2024occllamaoccupancylanguageactiongenerativeworld} model autonomous driving planning tasks based on occupancy world models.

\subsection{Temporal Fusion in Autonomous Driving}
\vspace{-0.5ex}
Temporal fusion has attracted growing interest in autonomous driving~\cite{hu2022st,yang2023bevformer,liu2022petrv2,huang2022bevdet4d,park2022time}. In 3D object detection, BEVDet4D~\cite{huang2022bevdet4d}, BEVFormer v2 \cite{yang2023bevformer}, PETR v2 \cite{liu2022petrv2}, and StreamPETR~\cite{wang2023exploring} build upon BEVDet~\cite{huang2021bevdet}, BEVFormer~\cite{li2022bevformer}, and PETR~\cite{liu2022petr}, by integrating historical data, effectively improving detection performance.
SOLOFusion~\cite{park2022time} first explores long-term temporal fusion in BEV space. StreamPETR~\cite{wang2023exploring} and Sparse4D~v2~\cite{lin2023sparse4d} further advance temporal fusion by propagating object-centric temporal data via sparse queries across frames. VideoBEV~\cite{han2024exploring} uses recurrent fusion in BEV space for efficiency. Recent efforts extend temporal fusion to occupancy prediction. FB-Occ \cite{li2023fbocc}, COTR \cite{ma2023cotr}, SparseOcc \cite{liu2023fully}, OPUS \cite{wang2024opus}, ALOcc \cite{chen2024alocc}, \etc direct adapt temporal fusion techniques from 3D object detection to fuse volume features. OccWorld~\cite{zheng2025occworld}, OccSora~\cite{wang2024occsora}, ALOcc~\cite{chen2024alocc} leverage temporal information to predict future frames or occupancy flow. However, these methods only focus on voxel-level temporal fusion.
In contrast, this work thoroughly explores temporal dynamics in the VisionOcc framework and proposes a unified method to integrate diverse temporal representations.

\subsection{Optimization-Inspired RNNs}

Optimization-inspired RNNs \cite{gregor2010learning,zheng2015conditional,putzky2017recurrent,sun2024learning,behrouz2024titans} model classic optimization processes with RNN structures. LISTA \cite{gregor2010learning} converts ISTA’s iterations into network layers. CRFasRNN \cite{zheng2015conditional} mimics CRF inference through RNN steps. RIM \cite{putzky2017recurrent} unfolds inverse problems into timesteps, refining updates with RNNs. Recent works such as TTT \cite{sun2024learning}, Titans~\cite{behrouz2024titans} frame language model updates as a local optimization problem solved by RNNs. Similarly, this work employs RNNs to fuse diverse temporal cues in VisionOcc. Unlike them, it reinterprets RNNs as optimization processes, delivering a generalized, interpretable approach to unify the fusion of heterogeneous temporal representations.

\begin{figure*}[t]
    \centering
    \setlength{\abovecaptionskip}{0pt}
    \includegraphics[width=1. \linewidth]{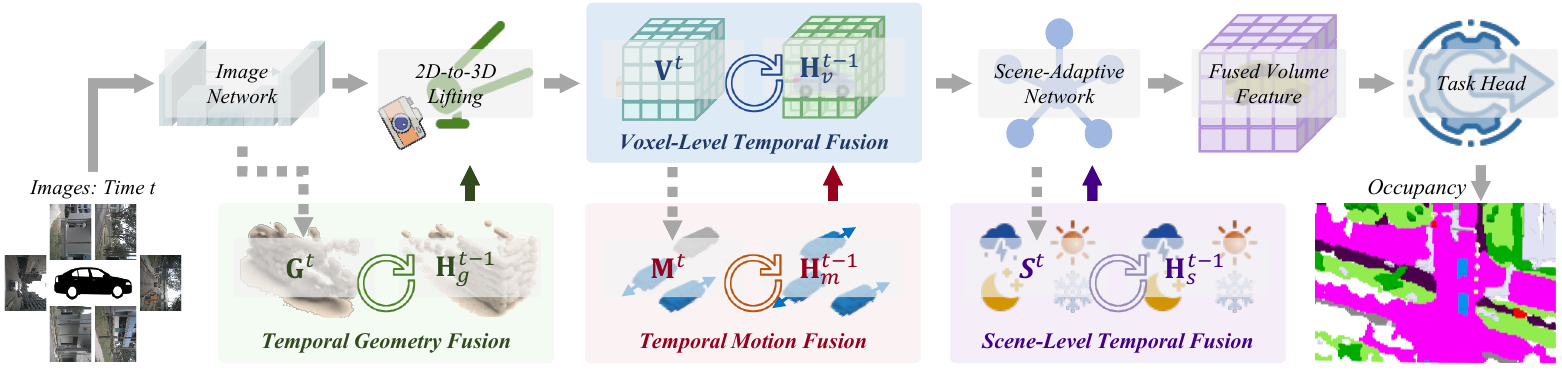}
    
    \caption{\textbf{Multi-level temporal fusion in the VisionOcc pipeline.} Volume features \(\mathbf{V}^t\), geometry \(\mathbf{G}^t\), motion \(\mathbf{M}^t\), and scene-adaptive parameters \(\mathbf{S}^t\) are enhanced through RNN-style temporal fusion, each capturing distinct temporal dynamics. Single-frame-sized historical states \(\mathbf{H}_v^{t-1}\), \(\mathbf{H}_g^{t-1}\), \(\mathbf{H}_m^{t-1}\), and \(\mathbf{H}_s^{t-1}\) are stored in memory and updated frame-by-frame.}
    
    \label{fig:framework}
    \vspace{-2mm}
\end{figure*}
\section{Multi-Level Temporal Cues in VisionOcc}
\label{sec:multi_level}
\subsection{VisionOcc Pipeline}
\label{sec:background}
VisionOcc processes a sequence of frames $\{I_i^{t}\}_{i=1,t=1}^{N,T}$ from $N$ camera views over $T$ timesteps. At each timestep $t$, the image network extracts 2D features $\{\mathbf{X}_i^{t}\}_{i=1}^{N}$ from each camera view. These multi-view features are transformed to 3D through a 2D-to-3D lifting function \cite{philion2020lift,li2022bevformer}:
\begin{equation}
    \mathbf{V}^t = \operatorname{Lift}(\{\mathbf{X}_i^{t}\}_{i=1}^{N}; \{\mathbf{G}_i^{t}\}_{i=1}^{N}),
\end{equation}
where $\operatorname{Lift}(\cdot;\cdot)$ denotes the lifting function, $\mathbf{G}^t$ represents the geometry prior (typically depth maps \cite{huang2022bevdet4d,li2023fbocc,chen2024alocc}), and $\mathbf{V}^t$ is the resulting 3D volume features. The task head $f_o$ further encodes and decodes these volume features to generate 3D semantic occupancy predictions, \ie,
\begin{equation}
    \mathbf{O}^t = f_o(\mathbf{V}^t) .
    \label{eq:f_o}
\end{equation}
Conventional VisionOcc approaches \cite{huang2022bevdet4d,li2023fbocc,chen2024alocc} typically use a fusion module to incorporate voxel-level temporal information before the task head. This module maintains a memory queue that stores volume features from the previous $N_h$ frames, \ie, $\mathbf{H}_v^{t-1} = \{\mathbf{V}^{t-1}, \mathbf{V}^{t-2}, ..., \mathbf{V}^{t-N_h}\}$, where $\mathbf{H}_v^{t-1}$ represents the historical memory state. 
When processing the current frame’s volume features, it fuses memorized temporal features to enhance them, \ie,
\begin{equation}
\vspace{-1ex}
    \mathbf{V}^t_f = \mathrm{Fuse}(\mathbf{V}^t, \mathbf{H}_v^{t-1}) .
\end{equation}
Here, $\mathbf{V}^t_f$ is the fused features, used to replace $\mathbf{V}^t$ in \cref{eq:f_o}.

\subsection{Temporal Cue Analysis and Formulation}
\label{sec:temp}
Traditional voxel-level temporal fusion \cite{li2023fbocc,ma2023cotr,chen2024alocc} stacks position-level evidence over time for reliable occupancy prediction. However, we argue that several temporal cues remain underutilized. Within the VisionOcc pipeline, we propose three distinct types of temporal information, each serving a unique role, as illustrated in \cref{fig:intro}.

\noindent \textbf{Scene-level Cue.}
The environment evolves continuously over short time spans, implying exploitable scene consistency priors. Accordingly, we design scene-level information to encode global properties, such as lighting, weather, and road characteristics. Drawing from domain adaptation \cite{ganin2015unsupervised,pmlr-v119-sun20b}, this information is expected to capture abstract, domain-specific features. It complements voxel-level features, which focus on per-voxel semantics. Inspired by test-time adaptation \cite{pmlr-v119-sun20b,ijcai2024p70,sun2024learning}, we introduce additional scene-adaptive network parameters $\mathbf{S}^t$ (\cref{sec:scene}). These parameters engage in inference across all volume features, capturing a global receptive field and serving as a proxy for scene information. They are dynamically updated with temporal inputs to enable continuous scene adaptation.

\noindent \textbf{Motion Cue.}
In voxel-level temporal fusion, historical features must align with the current frame’s coordinate system for positional semantic consistency. Ideally, this should handle dynamic object motion and camera pose estimation errors. We formulate the alignment with these corrections:
\begin{equation}
\scalebox{0.901}{$
\hat{\textbf{P}}\!=\!\mathbf{R}_{t\rightarrow t\!-\!1} (\mathbf{P}+\mathbf{M}^{t}),\  \mathrm{Warp}(\mathbf{H}_v^{t-\!1})\!=\!\mathrm{Sample}\! \left( \mathbf{H}_v^{t-\!1}; \hat{\textbf{P}} \right),
$}
\label{eq:sample}
\end{equation}
where $\textbf{P}$ is the standard volume grid coordinates, and $\textbf{R}_{t\rightarrow t-1}$ denotes the transformation matrix from the current to the last frame. The warping function $\mathrm{Warp}(\cdot)$ aligns features $\mathbf{H}_v^{t-\!1}$ to the current frame’s ego coordinate system. This is implemented via the $\mathrm{Sample}(\cdot;\cdot)$ function \cite{philion2020lift}, which samples features based on the transformed coordinates $\hat{\textbf{P}}$. The term $\mathbf{M}^{t}$ represents the offset intended to account for potential motion compensation and pose estimation error correction. It is estimated by a linear layer $f_m$:
\begin{equation}
    \mathbf{M}^{t} = f_m(\textbf{V}^t).
    \label{eq:m_t}
\end{equation}
Notably, the estimation of $\mathbf{M}^{t}$ requires no additional supervision. Instead, it is adaptively learned, similar to deformable networks \cite{dai2017deformable,zhu2020deformable}. As this learned offset facilitates positional alignment between frames, we refer to it as motion in the subsequent discussions, despite its implicit learning. Temporal coherence in consecutive frames allows historical motion patterns to refine current estimates. Details of the temporal motion fusion are provided in \cref{sec:motion}.

\noindent \textbf{Geometry Cue.}
In VisionOcc, depth maps typically act as geometric priors for 2D-to-3D lifting \cite{huang2021bevdet,li2023fbocc,chen2024alocc}, encoded as discretized probability distributions over uniform bins:
\begin{equation}
    \mathbf{G}^t = \{P(d=d_k |\mathbf{X}^t)\}_{k=1}^K, \quad P(d |\mathbf{X}^t) = f_g(\mathbf{X}^t),
\end{equation}
where $d_k$ is the center of the $k$-th depth bin, $K$ is the total number of bins, and the network $f_g$ estimates depth from image features. The view index of 2D features $\mathbf{X}^t$ is omitted for notational simplicity. We argue that scene geometry remains consistent over time (assuming motion is ignored in its estimation). Geometry derived from past camera viewpoints provides complementary information for the current frame. Thus, we propose temporal geometry fusion to yield a more robust geometric prior, detailed in \cref{sec:geo}.

\section{Gradient Descent-based Temporal Fusion}
\label{sec:fusion}
The complex representations of each temporal cue inherently make their fusion non-trivial. To address this, we propose a unified RNN-style framework to streamline and standardize the integration of distinct temporal representations (\ie, $\mathbf{V}$, $\mathbf{S}$, $\mathbf{M}$, $\mathbf{G}$, as defined earlier). \cref{fig:framework} illustrates the VisionOcc pipeline with multi-level temporal fusion, detailed in Alg.~\ref{alg:fusion}’s pseudocode, where fused information is highlighted in red. For notational consistency, temporally enhanced features use subscript $f$ (\eg, $\mathbf{V}_f$), and historical states are denoted by $\mathbf{H}$, further distinguished by subscripts.
\subsection{Modeling RNN Dynamics via Gradient Descent}
\label{sec:theory}
\begin{figure}[t]
    \centering
    \setlength{\abovecaptionskip}{0pt}
    \includegraphics[width=1\linewidth]{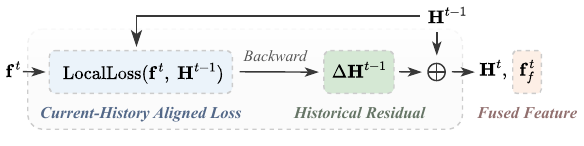}
    \caption{\textbf{Update dynamics of gradient descent-based temporal fusion pipeline.} $\mathbf{f}^t$ denotes the (geometry, motion, voxel-level, scene-level) feature of the current frame. $\mathbf{H}^{t-1}$ and $\mathbf{H}^{t}$ represent the prior and current historical states, respectively.}
    
    \label{fig:gradfusion}
\end{figure}

To unify temporal fusion, we reinterpret standard RNN updates as gradient descent steps to minimize discrepancies between current and historical information. This perspective allows us to integrate diverse temporal representation types within one theoretical framework. See \cref{fig:gradfusion} for an illustration of gradient-descent-based temporal fusion.

\begin{proposition}
    \label{prop:rnn}
    The RNN update step \( h^t = Ah^{t-1} + Bx^t \) is equivalent to a gradient descent step on \( h^{t-1} \) minimizing the loss function \( \mathcal{L}^t = \|Ah^{t-1} - Bx^t\|^2 \).
\end{proposition}

    \begin{proof}
    The gradient of \( \mathcal{L}^t \) \textit{\wrt} \( h^{t-1} \) is:
    \[
    \frac{\partial \mathcal{L}^t}{\partial h^{t-1}} = 2A^\top (Ah^{t-1} - Bx^t).
    \]
   Applying gradient descent with learning rate \(\eta\):
    \[
    h^t = h^{t-1} - \eta \cdot 2A^\top (Ah^{t-1} - Bx^t).
    \]
    Let \( A' = I - 2\eta A^\top A \) and \( B' = 2\eta A^\top B \), then:
    \[
    h^t = A' h^{t-1} + B' x^t.
    \]
    matching the RNN update form, proving the equivalence.
\end{proof}
\vspace{-1ex}
Next, we will detail how Prop.~\ref{prop:rnn} guides the fusion of distinct temporal cues, deriving loss functions and historical state update equations for each fusion type.

\subsection{Scene-Level Temporal Fusion}
\label{sec:scene}
We interpret scene-level fusion as a mini domain adaptation task over time. This perspective allows us to capture and adapt to varying environmental conditions across temporal sequences. Motivated by research in test-time adaptation \cite{pmlr-v119-sun20b,ijcai2024p70}, large language models (LLM) \cite{qin2022devil,sun2024learning}, and self-supervised learning \cite{chen2021exploring}, we propose encoding scene information into network parameters via a self-supervised task:
\begin{equation}
\mathcal{L}_s^t = \|f_s^t(\mathrm{aug_1}(\textbf{V}^t)) - \mathrm{aug_2}(\textbf{V}^t)\|^2 ,
\end{equation}
where \( \mathrm{aug}_1(\cdot) \) and \( \mathrm{aug}_2(\cdot) \) are distinct augmentation functions. Given the high dimensionality of \( \textbf{V}^t \), we use linear mappings as augmentations: \( \mathrm{aug}_i(\textbf{V}^t) = \mathbf{Q}_i \textbf{V}^t \), \( i = 1, 2 \), where \( \mathbf{Q}_i \in \mathbb{R}^{c \times c} \) are learnable parameters. With slight abuse of notation, \(\mathbf{V}^t\) henceforth denotes flattened volume features, \(c\) as the channel dimension, \(n\) as the flattened spatial dimensions. Inspired by sequence modeling in LLM  \cite{sun2024learning,behrouz2024titans}, we build the parameterized network $f_s$ as:
\begin{equation}
\scalebox{0.945}{$
f_s^t(\textbf{X};\textbf{S}^t) =\! \gamma^t \odot \hat{\mathbf{Z}}+\!\beta^t+\!\mathbf{X}, \  \hat{\mathbf{Z}}=\mathrm{Norm}(\mathbf{W}^t \mathbf{X} +\!\mathbf{b}^t),
$}
\end{equation}
where $\mathrm{Norm}(\cdot)$ denotes Z-score normalization, $\odot$ represents the Hadamard product, $\gamma^t \in \mathbb{R}^c$, $\beta^t \in \mathbb{R}^c$ are learnable scale and shift parameters for layer normalization, and $\mathbf{W}^t \in \mathbb{R}^{c \times c}$ and $\mathbf{b}^t \in \mathbb{R}^c$ form the linear transformation. These four parameters serve as an instantiation of scene-level temporal information, \ie,
\begin{equation}
\textbf{H}_s^t=\textbf{S}^t = \{\gamma^t, \beta^t, \mathbf{W}^t, \mathbf{b}^t\} ,
\end{equation}
where \(\mathbf{S}^0\) denotes the initial parameters. Building upon established meta-learning paradigms \cite{finn2017model,antoniou2018train,sun2024learning}, \(\mathbf{S}^0\) is optimized jointly with the network’s primary objective function during training. During inference, \(\mathbf{S}^t\) can be updated frame-by-frame from \(\mathbf{S}^0\), memorizing global scene information by minimizing $\mathcal{L}_s^t$. Following Prop. \ref{prop:rnn}, we compute the scene-level historical residual as:
\begin{equation}
\label{eq:nablas}
    \nabla_s^t =  \nabla_{\gamma^t, \beta^t, \mathbf{W}^t, \mathbf{b}^t} \mathcal{L}_s^t .
\end{equation}

This gradient is efficiently computed via matrix operations (see appendix):
    \begin{equation}
    \label{eq:nablas2}
        \scalebox{0.9}{$
        \begin{gathered}
            \nabla_\gamma^t = 2(\Delta_1 \odot \hat{\mathbf{Z}}^t)\mathbf{1}_n, \quad
        \nabla_\beta^t = 2\Delta_1\mathbf{1}_n, \\
    \nabla_W ^t = \Delta_3(\mathbf{Q}_1\mathbf{V}^t)^\top, \quad
        \nabla_\mathbf{b}^t = \Delta_3\mathbf{1}_n, \\
    \Delta_3 = \left(\Delta_2 - \frac{1}{n}\mathbf{1}_c\mathbf{1}_c^\top\Delta_2 - \frac{1}{n}\hat{\mathbf{Z}^t} \odot (\mathbf{1}_c\mathbf{1}_c^\top(\hat{\mathbf{Z}}^t \odot \Delta_2))\right) \oslash (\mathbf{1}_c\boldsymbol{\sigma}), \\
    \Delta_2 = 2\gamma^{t-1} \odot \Delta_1, \quad \Delta_1 = \gamma \odot  f_s^{t-1}(\mathbf{Q}_1\mathbf{V}^t) - \mathbf{Q}_2\mathbf{V}^t,
        \end{gathered}
        $}
    \end{equation}
where $\oslash$ denotes Hadamard division, $\mathbf{1}_c \in \mathbb{R}^{c \times 1}$ and $\mathbf{1}_n \in \mathbb{R}^{n \times 1}$ represent vectors of ones. To simplify notation, we define $\hat{\mathbf{Z}}^t=\text{Norm}(\mathbf{W}^{t-1} \mathbf{Q}_1\mathbf{V}^t +\mathbf{b}^{t-1})$ as the normalized features of $\mathbf{V}^t$, computed using parameters from time step $t-1$. The historical scene-level state $\mathbf{H}_s^t$ is updated as:
\begin{equation}
    \mathbf{H}_s^t = \mathbf{H}_s^{t-1} -\eta_s \nabla_s^t,\quad \textbf{S}_f^t=\mathbf{H}_s^t,
    \label{eq:h_s}
\end{equation}
where $\eta_s$ controls the update magnitude, and $\textbf{S}_f^t$ denotes the historical fused scene-level representation. Per specific scene, $\mathbf{H}_s^{t}$ undergoes continuous updates through the self-supervised task during both training and inference, enabling dynamic adaptation to changing scene conditions.

As shown in \cref{fig:framework}, the scene-level fusion module is positioned after voxel-level temporal fusion. At each time step $t$, we first update the scene-adaptive parameters via \cref{eq:h_s}, then perform forward propagation:

\begin{equation}
\hat{\mathbf{V}}^t = f_s^t(\mathbf{Q}_o\mathbf{V}_f^t;\textbf{S}_f^t),
\end{equation}
thus injecting historical scene-level information into the volume features. Here, $\mathbf{Q}_o$ denotes additional learnable weights, distinct from the augmentation matrices $\mathbf{Q}_1$, $\mathbf{Q}_2$.

\subsection{Temporal Motion Fusion}
\label{sec:motion}

To mitigate frame-to-frame alignment errors in voxel-level temporal fusion, we exploit historical motion information by minimizing the discrepancy between consecutive frames. This leads to our motion-based loss function:

\begin{equation}
\mathcal{L}_m^t = \left\| \mathrm{Warp}(\mathbf{H}_m^{t-1}) - \mathbf{M}^{t} \right\|^2 ,
\end{equation}
where $\mathbf{M}^{t}$ (\cref{eq:m_t}) denotes the predicted motion map for the current frame and $\mathbf{H}_m^{t-1}$ is the historical motion state. Same to \cref{eq:sample}, the warping operation is formulated as:

\begin{equation}
\scalebox{0.992}{$
    \mathrm{Warp}(\mathbf{H}_m^{t-1}) = \mathrm{Sample} \left( \mathbf{H}_m^{t-1}; \mathbf{R}_{t\rightarrow t\!-\!1} (\mathbf{P}\!+\!\mathbf{M}^{t}) \right) .
    $}
\end{equation}
The motion residual is computed as:
\begin{equation}
\label{eq:nablam}
\scalebox{0.9292}{$
    \nabla_m^{t} \!=\!\nabla_{\mathbf{M}^t}\mathcal{L}_m^t \!=\! 2 \left( \mathrm{Warp}(\mathbf{H}_m^{t-1})\! -\! \mathbf{M}^{t} \right) \!\left( \mathbf{R}_{t\rightarrow t\!-\!1}^\top \mathbf{J} \!-\! \mathbf{I} \right),\!
    $}
\end{equation}
where $\mathbf{J}$ represents the Jacobian matrix of the sampling operation \textit{\wrt} the sampling coordinates (detailed derivation in appendix). For notational simplicity, we omit spatial dimensions in this formulation. We use the derivative \textit{\wrt} $\mathbf{M}^{t}$ as the historical redidual, rather than $\mathbf{H}_m^{t-1}$, to avoid additional coordinate transformations.
The historical motion state is updated according to:
\begin{equation}
\mathbf{H}_m^{t} = \mathbf{M}^{t} - \eta_m  \nabla_m^{t} ,\quad \mathbf{M}^{t}_f=\mathbf{H}_m^{t}.
\label{eq:h_m}
\end{equation}
Here, $\eta_m$ balances current motion predictions against the warped historical motion. $\mathbf{M}^{t}_f$ is the fused motion, which not only gathers position-level clues from historical motion but also gets information from interpolation gradients (\cf deformable networks \cite{dai2017deformable,zhu2020deformable}). It is used for subsequent voxel-level temporal fusion, as shown in \cref{fig:framework}.

\begin{algorithm}[t]
\caption{Multi-Level Temporal Fusion in VisionOcc Pipeline at Time $t$}
\label{alg:fusion}
\begin{spacing}{1.09}
\begin{algorithmic}[1]

\Require Image features of time $t$: $\{\mathbf{X}_i^{t}\}_{i=1}^{N}$, historical states $\mathbf{H}_v^{t-1}$, $\mathbf{H}_s^{t-1}$, $\mathbf{H}_m^{t-1}$, $\mathbf{H}_g^{t-1}$, hyper-parameters $\eta_m$, $\eta_s$.
\Ensure Predicted 3D semantic occupancy map $\mathbf{O}^t$, updated historical states $\mathbf{H}_v^{t}$, $\mathbf{H}_s^{t}$, $\mathbf{H}_m^{t}$, $\mathbf{H}_g^{t}$.

\State \scalebox{0.928}{$\mathbf{G}^t \!=\! f_g(\mathbf{X})~\comment{Geometry~prediction}$}
\begin{center}
\vspace{-3mm}
\noindent\textcolor{green2}{\textit{\#\hspace{-0.0pt}\#\hspace{-0.0pt}\#} \textit{Temporal Geometry Fusion} \textit{\#\hspace{-0.0pt}\#\hspace{-0.0pt}\#}}
 \vspace{-3mm}
 \end{center}
\State \scalebox{0.928}{$\eta_g \!=\! \mathrm{sigmoid}\left(f_{\eta_g}\left(\mathrm{cat}(\mathrm{Warp_c}({\mathbf{H}}_g^{t-1}), \mathbf{G}^{t})\right)\right)~\comment{\crefcolor{gray}{eq:etag}}$}
\State \scalebox{0.928}{$\mathbf{H}_g^t \!=\! (1 - \eta_g) \hat{\mathbf{H}}_g^{t-1} + \eta_g \mathbf{G}^t, \quad \textcolor{red}{\mathbf{G}^{t}_f} = \mathbf{H}_g^{t}$ }

\State \scalebox{0.928}{$\mathbf{V}^t \!=\! \operatorname{Lift}(\{\mathbf{X}_i^{t}\}_{i=1}^{N}; \{\textcolor{red}{{\mathbf{G}_f}_i^{t}}\}_{i=1}^{N}) ~\comment{Lift~2D~to~3D}$}
\begin{center}
\vspace{-3mm}
\noindent\textcolor{green2}{\textit{\#\hspace{-0.0pt}\#\hspace{-0.0pt}\#} \textit{Temporal Motion Fusion} \textit{\#\hspace{-0.0pt}\#\hspace{-0.0pt}\#}}
 \vspace{-3mm}
 \end{center}
\State \scalebox{0.93}{$\mathbf{H}_m^t \!=\! \mathbf{M}^t - \eta_m \nabla_m^t, \  \textcolor{red}{\mathbf{M}^{t}_f} = \mathbf{H}_m^{t}~\comment{\nabla_\mathit{m}^\mathit{t}~is~from~\crefcolor{gray}{eq:nablam}}$}

\begin{center}
\vspace{-3mm}
\noindent\textcolor{green2}{\textit{\#\hspace{-0.0pt}\#\hspace{-0.0pt}\#} \textit{Voxel-Level Temporal Fusion} \textit{\#\hspace{-0.0pt}\#\hspace{-0.0pt}\#}}
  \vspace{-3mm}
 \end{center}
\State \scalebox{0.92}{$\mathbf{H}_v^{t} \!= \!A_v \mathrm{Sample}(\mathbf{H}_v^{t-1}; \!\mathbf{R}_{t\rightarrow t\!-\!1}\!(\mathbf{P}\!+\!\textcolor{red}{\mathbf{M}_f^{t}}\!)) \!\!+ \!\!B_v \mathbf{V}^t\!, \textcolor{red}{\mathbf{V}^t_f}\! =\! \mathbf{H}_v^t$}


\begin{center}
\vspace{-3mm}
\noindent\textcolor{green2}{\textit{\#\hspace{-0.0pt}\#\hspace{-0.0pt}\#} \textit{Scene-Level Temporal Fusion} \textit{\#\hspace{-0.0pt}\#\hspace{-0.0pt}\#}}
  \vspace{-3mm}
 \end{center}
\State \scalebox{0.928}{$\mathbf{H}_s^t\! =\! \mathbf{H}_s^{t-1} \!- \!\eta_s \nabla_s^t,\ \textcolor{red}{\mathbf{S}_f^t}\!=\!\mathbf{H}_s^t~\comment{\nabla_s^t~is~from~Eqs.~(\!\refcolor{gray}{eq:nablas}, \refcolor{gray}{eq:nablas2})}$}

\State \scalebox{0.928}{$\hat{\mathbf{V}}^t\! = \!f_s^t(\mathbf{Q}_o\textcolor{red}{\mathbf{V}_f^t};\textcolor{red}{\mathbf{S}_f^t})$}

\State \scalebox{0.928}{$\mathbf{O}^t \!= \!f_o(\hat{\mathbf{V}}^t)~\comment{Map~to~3D~semantic~occupancy}$ }

\State \Return $\mathbf{O}^t$, $\mathbf{H}_v^{t}$, $\mathbf{H}_s^{t}$, $\mathbf{H}_m^{t}$, $\mathbf{H}_g^{t}$
\end{algorithmic}
\end{spacing}
\end{algorithm}
\subsection{Temporal Geometry Fusion}
\label{sec:geo}
To further enhance the 2D-to-3D lifting process, we leverage historical geometry as prior information, introducing a geometry fusion loss function defined as follows:
\begin{equation}
\mathcal{L}_g^t = \left\| \mathrm{Warp}_c(\mathbf{H}_g^{t-1}) - \mathbf{G}^{t} \right\|^2 .
\end{equation}
Here, $\mathbf{G}^{t}$ represents the geometry representation of the current frame, while $\mathbf{H}_g^{t-1}$ symbolizes the historical geometry state. The function $\mathrm{Warp}_c(\cdot)$ transforms $\mathbf{H}_g^{t-1}$ from the previous camera coordinate system to the current one. For notational simplicity, we define $\hat{\mathbf{H}}_g^{t-1} = \mathrm{Warp}_c(\mathbf{H}_g^{t-1})$.
The historical geometry residual is then computed as:
\begin{equation}
\nabla_g^t = \nabla_{\hat{\mathbf{H}}_g^{t-1}} \mathcal{L}_g^t = 2\left(\hat{\mathbf{H}}_g^{t-1} - \mathbf{G}^{t}\right).
\label{eq:nabla_g}
\end{equation}
The updated rule of the historical geometric state is:
\begin{equation}
\mathbf{H}_g^{t} = \hat{\mathbf{H}}_g^{t-1} - \eta_g \nabla_g^t,
\end{equation}
where $\eta_g$ is a learning rate that controls the balance between the current geometry prediction and the historical geometric residual. Incorporating \cref{eq:nabla_g} and, for simplicity, absorbing the factor of 2 into the learning rate, we obtain:
\begin{equation}
    \mathbf{H}_g^{t} = (1 - \eta_g) \hat{\mathbf{H}}_g^{t-1} + \eta_g \mathbf{G}^{t}, \quad \mathbf{G}^{t}_f=\mathbf{H}_g^{t} ,
    \label{eq:h_g}
\end{equation}
where $\mathbf{G}^{t}_f$ denotes the fused geometry representation, which is used for subsequent 2D-to-3D lifting, as shown in \cref{fig:framework}. Since $\mathbf{G}^{t}$ is predicted as discrete probabilities along the camera’s line of sight, we seek to preserve its normalized probabilistic distribution. To this end, we reinterpret its update as weighted averaging with a per-frame adaptive $\eta_g$, resembling GRU gating \cite{cho2014learning}.
Specifically, we define:
\begin{equation}
\label{eq:etag}
\eta_g = \mathrm{sigmoid}\left(f_{\eta_g}\left(\mathrm{cat}(\hat{\mathbf{H}}_g^{t-1}, \mathbf{G}^{t})\right)\right) ,
\end{equation}
where $\mathrm{cat}(\cdot)$ denotes vector concatenation, and the network $f_{\eta_g}$ maps the concatenated features to a scalar value.

\subsection{Memory\kern-0.04em-\kern-0.04emEfficient Voxel\kern-0.04em-\kern-0.04emLevel Temporal Fusion}
Traditional voxel-level temporal fusion \cite{huang2022bevdet4d,ma2023cotr,li2023fbocc,chen2024alocc} struggles with storing volume features from \(N_h\) frames, causing excessive memory use as \(N_h\) grows. To overcome this, we build on the central insight of this paper, employing a single-frame-sized RNN hidden state for temporal modeling. This is modeled directly using a vanilla RNN as:
\begin{equation}
\mathbf{H}_v^t = A_v \mathrm{warp}(\mathbf{H}_v^{t-1}) + B_v \mathbf{V}^t, \quad \mathbf{V}^t_f = \mathbf{H}_v^t .
\label{eq:vox_his}
\end{equation}
Here, $\mathbf{H}_v^{t}$ is redefined as the single-frame-sized historical voxel-level state, $\mathbf{V}^t_f$ is the temporal-enhanced volume feature, and $A_v, B_v \in \mathbb{R}^{c \times c}$ are learnable fusion parameters.

\subsection{Training and Inference with GDFusion}
GDFusion is a purely temporal fusion framework, similar to works such as SOLOFusion~\cite{park2022time}, StreamPETR~\cite{wang2023exploring}, and VideoBEV~\cite{han2024exploring}. Although its intermediate derivation involves loss functions and gradient descent, these serve solely as procedural steps in formulating our approach and \textbf{do not require any additional supervision}. During both training and inference, four temporal representations and their historical latent states are updated via \cref{eq:h_s,eq:h_m,eq:h_g,eq:vox_his}. This integrates historical information into the network in a streaming manner~\cite{han2024exploring,wang2023exploring,yuan2024streammapnet}. Notably, our method is agnostic to specific approaches and can be seamlessly adapted to various LSS-based VisionOcc techniques, as substantiated in \cref{sec:experiment}. The detailed VisionOcc pipeline incorporating GDFusion is presented in Alg.~\ref{alg:fusion}.

\section{Experiment}
\label{sec:experiment}
\textbf{Benchmarks.} 
Our experiments are based on the nuScenes dataset~\cite{caesar2020nuscenes}, which provides extensive data to develop and evaluate essential 3D perception algorithms.
This dataset comprises 1,000 scenes in total, with 700 designated for training, 150 for validation, and 150 for testing.
Occ3D~\cite{tian2024occ3d}, SurroundOcc~\cite{wei2023surroundocc}, and OpenOccupancy~\cite{wang2023openoccupancy} extend nuScenes with voxel-wise annotations at 0.4m, 0.5m, or 0.2m resolutions, respectively. Their spatial ranges are: Occ3D ($-40$m to $40$m in X/Y, $-1$m to $5.4$m in Z), SurroundOcc ($-50$m to $50$m in X/Y, $-5$m to $3$m in Z), and OpenOccupancy ($-51.2$m to $51.2$m in X/Y, $-5$m to $3$m in Z). Occ3D includes 18 semantic categories (17 object classes plus an \textit{empty} class), while SurroundOcc and OpenOccupancy exclude the \textit{others} class. As per prior tradition~\cite{wei2023surroundocc,wang2023openoccupancy,li2023fbocc,ma2023cotr,chen2024alocc}, experimenting on any single dataset is sufficient, but we tested all benchmarks for broader method comparisons. We use mIoU across all semantic classes as the primary metric. We also report the mIoU for dynamic object categories (mIoU\textsubscript{D})~\cite{chen2024alocc} and category-agnostic IoU as supplementary metrics, focusing on foreground and scene-overall performance, respectively.

\begin{figure}[t]
\centering
\setlength{\abovecaptionskip}{0pt}
\includegraphics[width=0.92\linewidth]{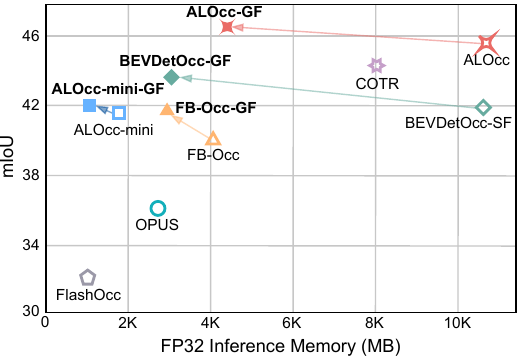}
\caption{\textbf{Inference memory consumption \vs occupancy prediction performance on the Occ3D benchmark.} Empty shapes denote baseline methods, while solid shapes represent GDFusion-enhanced versions. GDFusion significantly improves mIoU while reducing memory consumption.}
\label{fig:memory_sota}
\vspace{-1mm}
\end{figure}

\begin{figure}[t]
\centering
\setlength{\abovecaptionskip}{0pt}
\includegraphics[width=1\linewidth]{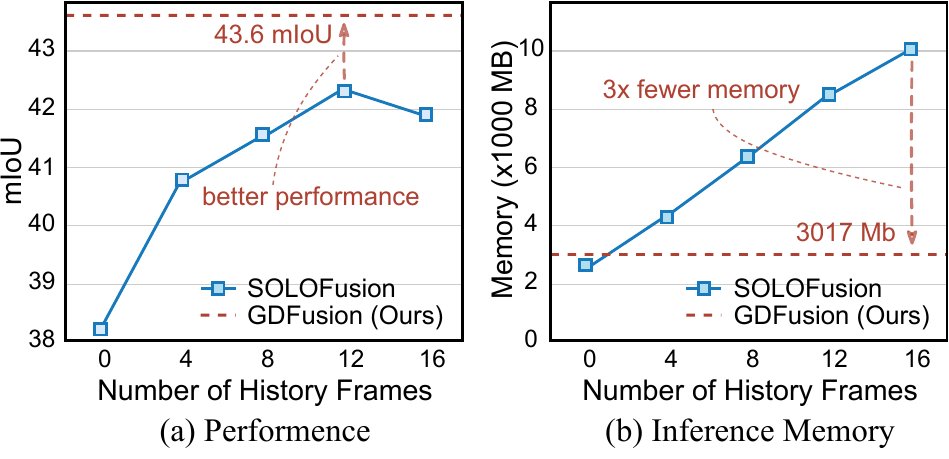}
\caption{\textbf{Comparison of our GDFusion and SOLOFusion \textit{\wrt} memory consumption}. SOLOFusion boosts performance with longer sequences but increases inference memory, while GDFusion achieves high performance with low memory consumption.}
\label{fig:memory}
\vspace{-1mm}
\end{figure}

\begin{table*}[t]

    \setlength{\abovecaptionskip}{0pt}
    \setlength{\tabcolsep}{0.015\linewidth}
    \begin{center}
        \resizebox{0.75\textwidth}{!}{
        \begin{tabular}{l|c|c|c|c|c|c|c}
            \toprule
            Method  & Backbone & Input Size  & mIoU & mIoU\textsubscript{D}  & IoU &FPS & Memory \\
            \midrule
            \midrule
            
            UniOCC~\cite{pan2023uniocc}  & ResNet-50 & $256\times704$  & 39.7 &- & -& - & - \\

            SurroundSDF~\cite{liu2024surroundsdf}  & ResNet-50 & $256\times704$  & 42.4 &36.2 & -& - & - \\
            FlashOCC~\cite{yu2023flashocc}  & ResNet-50 & $256\times704$ & 32.0 &24.7 & 65.3 &29.6 & 1205 \\
            COTR~\cite{ma2023cotr}  & ResNet-50 & $256\times704$  & 44.5 &38.6 & 75.0 &0.5 & 8063 \\
            ViewFormer~\cite{li2024viewformer}& ResNet-50 & $256\times704$   & 41.9 &35.0  & 70.2 &-& 2875 \\
            OPUS~\cite{wang2024opus}  & ResNet-50 & $256\times704$  &36.2 &33.3& 54.0 &8.2 & 2671 \\
            \midrule
            BEVDetOcc-SF~\cite{huang2022bevdet4d, park2022time}  & ResNet-50 & $256\times704$ & 41.9&34.4 & 75.1 &6.5 & 10717 \\
            \rowcolor{pink!10} \textbf{BEVDetOcc-GF} & ResNet-50 & $256\times704$  & \textbf{43.6{\color{red2}~$\uparrow$1.7}}&  \textbf{36.0{\color{red2}~$\uparrow$1.6}} & \textbf{77.8{\color{red2}~$\uparrow$2.7}}&7.0{\color{red2}~$\uparrow$0.5} & 3017{\color{red2}~$\downarrow$7700} \\
            \midrule
            FB-Occ~\cite{li2023fbocc}  & ResNet-50 & $256\times704$  & 39.8 &34.2 & 69.9 &10.3 & 4099 \\
            \rowcolor{pink!10} \textbf{FB-Occ-GF} & ResNet-50 & $256\times704$   &  \textbf{41.7{\color{red2}~$\uparrow$1.9}} & \textbf{35.8{\color{red2}~$\uparrow$1.6}} & \textbf{73.2{\color{red2}~$\uparrow$3.3}} &10.3& 2879{\color{red2}~$\downarrow$1224} \\
            \midrule
            ALOcc-mini \cite{chen2024alocc}  & ResNet-50 & $256\times704$  &41.4 &35.4  &70.0 &30.5 & 1605 \\
            \rowcolor{pink!10} \textbf{ALOcc-mini-GF}   & ResNet-50 & $256\times704$  & \textbf{42.0{\color{red2}~$\uparrow$0.6}} & \textbf{35.7{\color{red2}~$\uparrow$0.3}}& \textbf{71.9{\color{red2}~$\uparrow$1.9}} & 23.5{\color{blue}~$\downarrow$7} & 1179{\color{red2}~$\downarrow$426}\\
            \midrule
            ALOcc \cite{chen2024alocc}  & ResNet-50 & $256\times704$  &45.5 &39.3 & 75.3 &6.0 & 10793 \\
            \rowcolor{pink!10} \textbf{ALOcc-GF}   & ResNet-50 & $256\times704$  & \textbf{46.5{\color{red2}~$\uparrow$1.0}} & \textbf{40.2{\color{red2}~$\uparrow$0.9}}& \textbf{77.4{\color{red2}~$\uparrow$2.1}} &6.2{\color{red2}~$\uparrow$0.2}& 4347{\color{red2}~$\downarrow$6446}\\
            \midrule
            \midrule
            BEVFormer~\cite{li2022bevformer}  & ResNet-101 & $900\times1600$  & 39.2 &37.2  &-&4.4 &4667\\
            OSP~\cite{shi2024occupancy}& ResNet-101 & $900\times1600$   & 41.2 &37.0&-&-&2671\\
            FastOcc~\cite{hou2024fastocc}  & ResNet-101 & $640\times1600$ &34.5 &39.2  &-&-&-\\
            PanoOcc~\cite{wang2024panoocc}  & ResNet-101 & $640\times1600$  & 42.1&37.3 &- &-&-\\
            SurroundOcc~\cite{wei2023surroundocc}  & InternImage-Base  &$900\times1600$   & 40.7&- & -&-&-\\
           COTR~\cite{ma2023cotr}  & Swin-Base &$512\times1408$   &46.2&41.3&74.9  &-&-\\
           \midrule
            BEVDetOcc-SF  & Swin-Base    &$512\times1408$ &47.1  & 41.2&76.8  &2.2& 13682 \\
            \rowcolor{pink!10} \textbf{BEVDetOcc-GF} & Swin-Base  &$512\times1408$   &  \textbf{48.5{\color{red2}~$\uparrow$1.4}} & \textbf{42.9{\color{red2}~$\uparrow$1.8}} & \textbf{79.5{\color{red2}~$\uparrow$2.7}} &2.2& 4492{\color{red2}~$\downarrow$9190} \\
            \midrule
            ALOcc   & Swin-Base  &$512\times1408$  &50.6 &46.1 & 78.1 &1.5 &24072  \\
            \rowcolor{pink!10} \textbf{ALOcc-GF}    & Swin-Base  &$512\times1408$    & \textbf{51.8{\color{red2}~$\uparrow$1.2}} & \textbf{47.4{\color{red2}~$\uparrow$1.3}}& \textbf{79.9{\color{red2}~$\uparrow$1.8}} &1.8{\color{red2}~$\uparrow$0.3} & 8312{\color{red2}~$\downarrow$15760}\\
            \bottomrule
        \end{tabular}
        }
    \end{center}
      \vspace{-1.5mm}
        \caption{\textbf{Comparison of 3D semantic occupancy prediction performance on the Occ3D benchmark, evaluated with mIoU, mIoU\textsubscript{D}, and IoU metrics.} Relative improvements are highlighted with red arrows {\color{red2}~$\uparrow$}. FPS for FlashOCC, COTR, OPUS, and BEVFormer are sourced from \cite{chen2024alocc}, evaluated on NVIDIA RTX 4090. Other FPS and memory values are our evaluations on RTX 4090. Our GDFusion (\textbf{-GF}) consistently enhances the performance of existing 3D semantic occupancy prediction methods and reduces inference memory consumption.}
   
   \label{tab:sota}
\end{table*}

\begin{table*}[t]

    \setlength{\abovecaptionskip}{0pt}
    \setlength{\tabcolsep}{0.015\linewidth}
    \begin{center}
        \resizebox{0.75\textwidth}{!}{
        \begin{tabular}{l|c|c|c|c|c|c|c}
            \toprule
            Method  & Backbone & Input Size  & mIoU & mIoU\textsubscript{D}  & IoU &FPS & Memory \\
            \midrule
            \midrule
            
            BEVFormer~\cite{li2022bevformer}  & ResNet-101 & $900\times1600$  & 16.8 &14.2 & 30.5 &3.3 & 25100 \\

            TPVFormer~\cite{huang2023tri}  & ResNet-101 & $900\times1600$  & 17.1 &14.0 & 30.9 &2.9 & 29000 \\
            SurroundOcc~\cite{wei2023surroundocc}  & ResNet-101 & $900\times1600$ & 20.3 &18.4 & 31.5&- &-  \\

            GaussianFormer~\cite{huang2024gaussianformer}  & ResNet-101 & $900\times1600$  & 19.1 &17.3 & 29.8 &2.7 &6229  \\
            
            GaussianWorld~\cite{zuo2024gaussianworld}  & ResNet-101 & $900\times1600$  & 22.1 &19.7 & 33.4 &4.4 &7030  \\
            
            \midrule
            BEVDetOcc~\cite{huang2022bevdet4d, park2022time}  & ResNet-50 & $900\times1600$ & 17.5&14.1 & 29.2  &1.1 & 8191 \\
            \rowcolor{pink!10} \textbf{BEVDetOcc-GF} & ResNet-50 &  $900\times1600$  
            & \textbf{20.8{\color{red2}~$\uparrow$3.3}}&  \textbf{16.2{\color{red2}~$\uparrow$2.1}} & \textbf{35.3{\color{red2}~$\uparrow$6.1}} & 1.1 & 8447{\color{blue}~$\uparrow$256} \\
            \midrule
            ALOcc-mini*~\cite{chen2024alocc}  & ResNet-50 & $900\times1600$  & 21.5 &19.5 &31.5  &5.8& 2869 \\
            \rowcolor{pink!10} \textbf{ALOcc-mini-GF} & ResNet-50 &$900\times1600$   &  \textbf{23.1{\color{red2}~$\uparrow$1.6}} & \textbf{20.3{\color{red2}~$\uparrow$0.8}} & \textbf{34.6{\color{red2}~$\uparrow$3.1}}  &5.4{\color{blue}~$\downarrow$0.4}& 3441{\color{blue}~$\uparrow$572} \\
            \midrule
            ALOcc* \cite{chen2024alocc}  & ResNet-50 & $900\times1600$  &24.0 &21.7 & 34.7 &1.0 & 11117 \\
            \rowcolor{pink!10} \textbf{ALOcc-GF}   & ResNet-50 &$900\times1600$ & \textbf{25.5{\color{red2}~$\uparrow$1.5}} & \textbf{22.5{\color{red2}~$\uparrow$0.8}}& \textbf{38.2{\color{red2}~$\uparrow$3.5}}  &0.9{\color{blue}~$\downarrow$0.1} & 11857{\color{blue}~$\uparrow$740}\\
            
            \bottomrule
        \end{tabular}
        }
    \end{center}
      \vspace{-1.5mm}
        \caption{\textbf{Comparison of 3D semantic occupancy prediction performance on the SurroundOcc benchmark.} FPS and memory for BEVFormer and TPVFormer are sourced from \cite{lu2023octreeocc}, evaluated on NVIDIA A100. FPS and memory for GaussianFormer and GaussianWorld are obtained from their respective papers, evaluated on NVIDIA RTX 4090. All other models are evaluated by us on NVIDIA RTX 4090. * denotes versions of the models with temporal fusion removed. GDFusion adds negligible inference time and memory overhead.}
   
   \label{tab:sota_surroundocc}
    \vspace{-1mm}
\end{table*}

\begin{table*}[t]
    \setlength{\abovecaptionskip}{0pt}
    \setlength{\tabcolsep}{0.015\linewidth}
    \begin{center}
        \resizebox{0.77\textwidth}{!}{
        \begin{tabular}{l|c|c|c|c|c|c|c|c}
            \toprule
            Method &Input & Backbone & Input Img. Size  & mIoU & mIoU\textsubscript{D}  & IoU &FPS & Memory \\
            \midrule
            \midrule
        
            TPVFormer \cite{huang2023tri} &C & ResNet-50 & $900\times1600$ &7.8 & 11.7 &15.3 &-&- \\
             C-CONet \cite{wang2023openoccupancy} &C & ResNet-50 & $900\times1600$  &12.8 &10.6 & 20.1 &3.5& - \\
              C-OccGen \cite{wang2024occgen} &C & ResNet-50 & $900\times1600$ &14.5 & 11.7 &23.4 &3.0&- \\

            \midrule
            ALOcc-2D* \cite{chen2024alocc} &C & ResNet-50 & $900\times1600$  &15.7 &11.5 & 25.2 &0.8& 13805 \\
            \rowcolor{pink!10} \textbf{ALOcc-2D-GF} &C  & ResNet-50 &$900\times1600$ & \textbf{17.9{\color{red2}~$\uparrow$2.2}} & \textbf{13.7{\color{red2}~$\uparrow$2.2}}& \textbf{28.6{\color{red2}~$\uparrow$3.4}}  &0.8 &13857{\color{blue}~$\downarrow$52}\\
 \midrule
            \midrule
            L-CONet \cite{wang2023openoccupancy} &L &- & -  &15.8 &10.8 & 30.9 &4.0& - \\
            L-OccGen \cite{wang2024occgen} &L & - & -  &16.8 &11.8 & 31.6 &3.7& - \\
            3DSketch$^{\dagger}$ \cite{chen20203d}  &C\&D & ResNet-50 & $900\times1600$  &10.7 &7.4 & 25.6 &-&-  \\
              AICNet$^{\dagger}$ \cite{li2020anisotropic}  &C\&D & ResNet-50 & $900\times1600$  &10.6 &7.4 & 23.8 &-&-  \\
            M-CONet \cite{wang2023openoccupancy} &C\&L &ResNet-50 &$900\times1600$  &20.1 &18 & 29.5 &2.9&-  \\
             M-OccGen \cite{wang2024occgen} &C\&L &ResNet-50 &$900\times1600$  &22.0 &20 &30.3 &2.3& - \\
              Co-Occ \cite{pan2024co} &C\&L &ResNet-101 &$900\times1600$  &21.9 &19.5 &30.6 &-& - \\
             OccLoff \cite{zhang2025occloff} &C\&L &ResNet-101 &$900\times1600$  &22.9 &21.1 &31.4 &3.5 & 8089 \\
             
             \midrule
             
            ALOcc-2D* \cite{chen2024alocc}  &C\&D & ResNet-50 & $900\times1600$  &22.4 &20.3 & 30.4 & 0.8&13839  \\
            
            \rowcolor{pink!10} \textbf{ALOcc-2D-GF}   &C\&D & ResNet-50 &$900\times1600$ & \textbf{24.5{\color{red2}~$\uparrow$2.1}} & \textbf{21.6{\color{red2}~$\uparrow$1.3}}& \textbf{34.5{\color{red2}~$\uparrow$4.1}}  &0.8& 13891{\color{blue}~$\downarrow$52}\\
            
            \bottomrule
        \end{tabular}
        }
    \end{center}
      \vspace{-1.5mm}
        \caption{\textbf{Comparison of 3D semantic occupancy prediction performance on the OpenOccupancy benchmark.} C, L, D denote \textit{camera}, \textit{LiDAR}, and \textit{depth} inputs. $\dagger$ denotes models that concatenate individual camera view outputs for surrounding results. FPS and memory for ALOcc-2D* and ALOcc-2D-GF are evaluated by us on NVIDIA RTX 4090. OccLoff values are from its paper (NVIDIA A100). Other models’ FPS and memory are cited from \cite{wang2024occgen} (NVIDIA V100). In the last two rows, LiDAR-derived depth is used during 2D-to-3D lifting to extend ALOcc-2D into a multimodal version for comprehensive comparisons.}
   
   \label{tab:sota_openoccupancy}
\end{table*}
\noindent \textbf{Implementation Details.} 
We evaluate GDFusion by incorporating it into $5$ baseline models, BEVDetOcc \cite{huang2022bevdet4d}, FB-Occ \cite{li2023fbocc}, ALOcc, ALOcc-mini, and ALOcc-2D \cite{chen2024alocc}, yielding the enhanced variants BEVDetOcc-GF, FB-Occ-GF, ALOcc-GF, ALOcc-mini-GF, and ALOcc-2D-GF. Following prior work~\cite{li2024viewformer,chen2024alocc,yu2023flashocc}, we use a ResNet-50~\cite{resnet} backbone with $256 \times 704$ image inputs for most Occ3D experiments, scaling to a Swin Transformer Base~\cite{liu2021Swin} backbone with $512 \times 1408$ image inputs for larger-scale tests. For SurroundOcc and OpenOccupancy, we adopt a ResNet-50 backbone with $900 \times 1600$ image inputs. Similar to FB-Occ and ALOcc, we adopt the video-level data shuffling strategy from SOLOFusion \cite{park2022time} during training. For BEVDet, we follow \cite{li2023fbocc,chen2024alocc} to implement a 16-frame temporal fusion version based on SOLOFusion as a baseline (BEVDetOcc-SF). Both the baselines and the versions with GDFusion were trained for 12 epochs using CBGS \cite{park2022time} with a learning rate of $2 \times 10^{-4}$ and a global batch size of 16. The image and BEV augmentation strategies are the same as in the original BEVDet codebase \cite{huang2022bevdet4d}. Relevant parameters for FB-Occ and ALOcc are kept the same as in their original papers.

\begin{table}[t]
        \centering
        \small
        \setlength{\tabcolsep}{4pt}
        \resizebox{1.\linewidth}{!}{
        \begin{tabular}{c|l|ccc}
            \toprule
          Exp. & Configuration  & mIoU &  mIoU\textsubscript{D} &  IoU  \\
            \midrule
            \midrule
         0& Baseline  & 38.0 & 31.0& 71.1  \\
         1& + SOLOFusion (16f)  & 41.4 & 34.0 &74.4 \\
         2&+ History fusion from start &41.2 & 33.6 & 74.6 \\
            \midrule
           3& Baseline + RNN fusion &41.3 & 33.6& 74.8 \\
           4&+ History fusion from start& 41.4{\color{red2}~$\uparrow$0.1}& 33.9{\color{red2}~$\uparrow$0.3} & 74.8{\color{red2}~$\uparrow$0}  \\
           5&+ No video split &41.6{\color{red2}~$\uparrow$0.2}  & 33.7{\color{blue}~$\downarrow$0.2}& 75.3{\color{red2}~$\uparrow$0.5} \\
          \rowcolor{pink!10}  6&+ Sinusoidal time embedding &41.8{\color{red2}~$\uparrow$0.2} & 34.0{\color{red2}~$\uparrow$0.3}& 76.5{\color{red2}~$\uparrow$1.2}  \\
            \bottomrule
        \end{tabular}
        }
          \vspace{-2mm}
         \caption{\textbf{Ablations on improving RNN-based voxel-level fusion.}}
       
        \label{tab:voxhis}
    \end{table}

\subsection{Memory Consumption}
GDFusion reduces inference memory by storing all historical information in a single-frame-sized hidden state. As shown in \cref{tab:sota} and \cref{fig:memory_sota}, replacing baseline temporal fusion modules with GDFusion significantly reduces memory consumption. As shown in \cref{tab:sota_surroundocc,tab:sota_openoccupancy}, GDFusion introduces negligible memory overhead compared to baselines without temporal fusion. To emphasize our approach's memory efficiency, we further compared it with SOLOFusion on Occ3D, using BEVDetOcc as the baseline. As illustrated in \cref{fig:memory}, the memory consumption of SOLOFusion scales linearly with the number of stored historical frames, whereas GDFusion maintains memory usage comparable to non-temporal fusion levels. Despite utilizing less memory, GDFusion harnesses information from all historical frames and outperforms the long-history fusion variant of SOLOFusion. Notably, although GDFusion stores various types of temporal information, the additional memory overhead relative to voxel features is negligible.

\subsection{Comparison with SOTAs}
\cref{tab:sota,tab:sota_surroundocc,tab:sota_openoccupancy} compare GDFusion with SOTA methods. We reimplemented 5 baseline models (BEVDetOcc, FB-Occ, ALOcc, ALOcc-mini, ALOcc-2D) using official codes, citing or reevaluating other methods with official checkpoints. 

On Occ3D~(\cref{tab:sota}), we primarily compare GDFusion against the original temporal fusion schemes of baselines. GDFusion consistently delivers substantial improvements across all metrics: 1.4\%-4.8\% gains in mIoU (all-class average), 0.8\%-4.7\% in mIoU\textsubscript{D} (dynamic object classes), and remarkable 1.8-3.3 point increases in IoU, reflecting superior scene geometry quality. Additionally, GDFusion reduces memory usage by 27\%-72\% compared to the originals by leveraging comprehensive temporal data in a streaming manner. In terms of inference speed, GDFusion outperforms multi-frame stacking temporal fusion schemes in most cases, except for ALOcc-mini-GF, where smaller BEV feature dimensions limit speed advantages.

On SurroundOcc~(\cref{tab:sota_surroundocc}) and OpenOccupancy~(\cref{tab:sota_openoccupancy}), we compare baselines without temporal fusion against GDFusion-enhanced versions. We omit FB-Occ results on SurroundOcc due to an unexpected bug in its implementation. On OpenOccupancy, the large occupancy size made vanilla BEVDetOcc, FB-Occ, and ALOcc incompatible, so we report ALOcc-2D results. GDFusion significantly enhances performance across these benchmarks with negligible computational and memory overhead. Notably, on OpenOccupancy, where larger image inputs and occupancy sizes are used, GDFusion’s additional cost is nearly immeasurable relative to baseline inference.

\begin{table}
 \centering
        \setlength{\tabcolsep}{5pt}
        \resizebox{0.99\linewidth}{!}{
        \begin{tabular}{l|cccc|ccc}
            \toprule
             Exp. & Voxel & Scene & Motion & Geometry & mIoU & mIoU\textsubscript{D} & IoU  \\
            \midrule
            \midrule
             B & & & & & 38.0 &31.0 & 71.1 \\
             BG & & & & \checkmark & 39.2&32.0 & 73.0 \\
             BS & & \checkmark & & & 39.0 &31.9 & 74.1 \\
             BV & \checkmark & & & & 41.8 &34.0& 76.5 \\
             BVS & \checkmark & \checkmark & & & 42.5 &34.8& 77.0 \\
             BVM& \checkmark &  & \checkmark & & 42.4 &34.9& 76.4 \\
             BVG & \checkmark & & & \checkmark & 42.7 &35.3& 73.8 \\
             BVMG & \checkmark & & \checkmark & \checkmark & 43.0 &\textbf{35.5}& 77.0 \\
             BVSG & \checkmark & \checkmark & & \checkmark & 43.0 &\textbf{35.5}& 77.5 \\
             BVSM & \checkmark & \checkmark & \checkmark &  & 42.6&34.8 & 77.3 \\
           \rowcolor{pink!10}  Full & \checkmark & \checkmark & \checkmark & \checkmark & \textbf{43.3} &35.3& \textbf{77.8} \\
            \bottomrule
        \end{tabular}}
         \vspace{-0.5mm}
         \caption{\textbf{Ablations on different temporal cues.} B, G, V, S, and M represent baseline, temporal geometry fusion, voxel-level fusion, scene-level fusion, and temporal motion fusion, respectively.}
        
        \vspace{-1mm}
       \label{tab:temporal}
\end{table}

\subsection{Ablation Study}
In this subsection, we conduct an ablation study on Occ3D. Unless stated otherwise, we use a spatially compressed BEVDet as our baseline model, which compresses height features into the channel dimension and applies 2D convolution for volume encoding \cite{yu2023flashocc}.

\noindent \textbf{Pre-Uplift~on~Voxel-Level~Temporal~Fusion.}~As shown in \cref{tab:voxhis}, vanilla RNN-based temporal fusion \textbf{(Exp.~3)} achieved comparable performance to SOLOFusion \textbf{(Exp.~1)}, demonstrating its effectiveness. We then explored several further refinements. First, the previous method \cite{li2023fbocc} delays fusion training until the second epoch to avoid noisy gradients from immature features (\textbf{Exps.~1-2}). \textbf{Exps.~3-4} show that RNN-based fusion does not require this delay, as its single-frame hidden state is less affected by erroneous gradients than multi-frame stacking methods. Second, SOLOFusion \cite{park2022time} splits videos to diversify training, whereas RNN fusion excels with continuous videos (\textbf{Exp.~5}), avoiding discontinuities and utilizing full temporal information. Finally, adding sinusoidal time embedding \cite{vaswani2017attention} further boosts performance (\textbf{Exp.~6}).

\begin{table}
\centering
 \setlength{\tabcolsep}{6pt}
    \resizebox{0.82\linewidth}{!}{
        \begin{tabular}{@{}lcc@{}}
        \toprule
        Module  & Runtime (ms) & Proportion (\%) \\ 
        \midrule
        \midrule
BEVDet-SF (16) & 155.6 & - \\
\midrule
\rowcolor{pink!10} BEVDet-GF (ours full) &146.6 & 100\\
Scene-AutoGrad & 8.8 & - \\
       Scene-Custom Matmul & 6.7 & 4.57 \\
       
        Motion-AutoGrad & 9.9 & - \\
      Motion-Custom Matmul  & 7.7 & 5.25 \\
       
         Geometry & 4.8 & 3.27 \\
        Voxel &6.5 & 4.43 \\

        \bottomrule
        \end{tabular}
        }
         \vspace{-1.mm}
     \caption{\textbf{Runtime analysis of key components, \vs SOLOFusion (16 frames).} AutoGrad refers to \textit{PyTorch}'s automatic differentiation; Custom Matmul is our custom matrix multiplication. Fusion modules are abbreviated (\eg, scene, voxel) for clarity. }
         
        \vspace{-1mm}
        \label{tab:runtime}
\end{table}

\noindent \textbf{Effect of Individual Temporal Cues.} 
The ablation results in \cref{tab:temporal} highlight the role of different temporal cues in our GDFusion framework. Using only the geometry cue \textbf{(BG)}, scene cue \textbf{(BS)}, or voxel-level temporal fusion \textbf{(BV)} each brought a substantial boost in performance. Combining different temporal cues \textbf{(BVS - Full)} led to further improvements, with motion \textbf{(BVM)} adding gains atop voxel-level fusion. Integrating all cues (voxel-level, scene-level, motion, and geometry) yielded optimal performance \textbf{(Full)}. These findings underscore the complementary nature of different temporal cues. Omitting any of them reduced accuracy, affirming the value of GDFusion’s comprehensive integration for robust 3D scene understanding.

\noindent \textbf{Wall-Clock Time.}
We conduct a runtime analysis of key components in GDFusion. As shown in \cref{tab:runtime}, our method outperforms the multi-frame stacking method SOLOFusion in total time consumption. Each temporal fusion component consumes less than 6\% of the total inference time (146.6ms), demonstrating the efficiency of GDFusion. Moreover, our custom matrix multiplication method for backpropagation outperforms \textit{PyTorch}'s built-in AutoGrad, achieving 32\% and 22\% efficiency gains in scene-level and motion fusion, respectively, demonstrating enhanced deployment feasibility.

\section{Conclusion and Discussion}
In this work, we introduce a novel temporal fusion framework for vision-based semantic occupancy prediction. Our method examines untapped temporal signals in VisionOcc, including scene-level, motion, and geometry cues. We reinterpret vanilla RNNs from a gradient descent perspective, unifying the fusion of diverse temporal representations. The proposed GradFusion integrates these signals seamlessly, boosting performance over baselines while significantly reducing memory usage. Our contributions include exploring varied temporal cues for VisionOcc and advancing gradient descent-based fusion techniques for complex temporal data. We believe the principles outlined here extend beyond VisionOcc, potentially benefiting a wide range of tasks.

{
    \small
    \bibliographystyle{ieeenat_fullname}
    \bibliography{main}
}
\clearpage

\onecolumn
\setcounter{page}{1}
\appendix
\renewcommand\thefigure{A.\arabic{figure}} 
\renewcommand\theequation{A.\arabic{equation}} 
\renewcommand\thetable{A.\arabic{table}} 

\section{Derivation of the Gradients}
\subsection{Gradient Computation for Scene-Level Temporal Fusion}

Consider a normalized transformation function $f_s: \mathbb{R}^{c \times N} \rightarrow \mathbb{R}^{c \times N}$ defined as:
\begin{equation}
    f_s(\mathbf{X}) = \gamma \odot \mathrm{Norm}(W\mathbf{X} + \mathbf{b}) + \beta + \mathbf{X},
\end{equation}
where $\gamma, \beta \in \mathbb{R}^{c \times 1}$ are learnable scale and shift parameters, $W \in \mathbb{R}^{c \times c}$ is a weight matrix, $\mathbf{b} \in \mathbb{R}^{c \times 1}$ is a bias vector, and $\odot$ denotes the Hadamard product. $\mathrm{Norm}()$ represents the Z-score normalization across the channel dimension.
We seek to minimize the loss function $\mathcal{L}_s$, defined as:
\begin{equation}
    \mathcal{L}_s = \|f_s(\mathbf{Q}_1\mathbf{V}^t) - \mathbf{Q}_2\mathbf{V}^t\|_F^2,
\end{equation}
where $\mathbf{Q}_1, \mathbf{Q}_2 \in \mathbb{R}^{c \times c}$ are given matrices and $\mathbf{V}^t \in \mathbb{R}^{c \times N}$. Here, $\|\cdot\|_F$ denotes the Frobenius norm.
To facilitate the gradient computation $\mathbf{H}_s^t = -\eta_s \nabla_{\gamma, \beta, W, \mathbf{b}} \mathcal{L}_s$, we introduce the following intermediary terms:
\begin{align}
    \Delta_1 &= f_s(\mathbf{Q}_1\mathbf{V}^t) - \mathbf{Q}_2\mathbf{V}^t, \\
    \mathbf{Z} &= W\mathbf{Q}_1\mathbf{V}^t + \mathbf{b}\mathbf{1}_n^\top,
\end{align}
where $\mathbf{1}_n \in \mathbb{R}^{n \times 1}$ denotes a vector of ones. The loss function can then be expressed as:
\begin{equation}
    \mathcal{L}_s = \frac{1}{nc}tr(\Delta_1^\top \Delta_1).
\end{equation}
Let $\hat{\mathbf{Z}} = \mathrm{Norm}(\mathbf{Z})$, we have:
\begin{align}
    \hat{\mathbf{Z}} &= (\mathbf{Z} - \boldsymbol{\mu}) \oslash \boldsymbol{\sigma}, \\
    \boldsymbol{\mu} &= \frac{1}{n} \mathbf{1}_c^\top \mathbf{Z}, \\
    \boldsymbol{\sigma}^2 &= \frac{1}{n} \mathbf{1}_c^\top \left( \mathbf{Z} - \mathbf{1}_c \boldsymbol{\mu} \right)^2,
\end{align}
where $\oslash$ denotes Hadamard division. The function $f_s$ can then be written as:
\begin{equation}
    f_s(\mathbf{Q}_1\mathbf{V}^t) = \gamma \odot \hat{\mathbf{Z}} + \beta + \mathbf{Q}_1\mathbf{V}^t.
\end{equation}
The gradient of the loss with respect to $f_s$ is:
\begin{equation}
    \frac{\partial \mathcal{L}_s}{\partial f_s} = 2\Delta_1.
\end{equation}
The gradients with respect to the learnable parameters $\gamma$ and $\beta$ are:
\begin{align}
    \frac{\partial \mathcal{L}_s}{\partial \gamma} &= 2(\Delta_1 \odot \hat{\mathbf{Z}})\mathbf{1}_n, \quad
    \frac{\partial \mathcal{L}_s}{\partial \beta} = 2\Delta_1\mathbf{1}_n.
\end{align}
For the normalized activations $\hat{\mathbf{Z}}$, we have:
\begin{equation}
    \frac{\partial \mathcal{L}_s}{\partial \hat{\mathbf{Z}}} = 2\gamma \odot \Delta_1 = \Delta_2,
\end{equation}
where we define $\Delta_2$ for notational convenience.
For each column $i$, the gradient of the normalized activations with respect to the pre-normalized activations is:
\begin{equation}
    \frac{\partial \hat{\mathbf{Z}}_i}{\partial \mathbf{Z}_i} = \frac{1}{\sigma_i}(\mathbf{I} - \frac{1}{n}\mathbf{1}_c\mathbf{1}_c^\top - \frac{1}{n}\hat{\mathbf{Z}}_i\hat{\mathbf{Z}}_i^\top),
\end{equation}
where $\mathbf{I}$ is the identity matrix.
The complete gradient with respect to $\mathbf{Z}$ is:
\begin{equation}
\scalebox{0.933}{$
    \frac{\partial \mathcal{L}_s}{\partial \mathbf{Z}} = \left(\Delta_2 - \frac{1}{n}\mathbf{1}_c\mathbf{1}_c^\top\Delta_2 - \frac{1}{n}\hat{\mathbf{Z}} \odot (\mathbf{1}_c\mathbf{1}_c^\top(\hat{\mathbf{Z}} \odot \Delta_2))\right) \oslash (\mathbf{1}_c\boldsymbol{\sigma}),
    $}
\end{equation}
Let $\Delta_3 = \frac{\partial \mathcal{L}_s}{\partial \mathbf{Z}}$, the gradients with respect to $W$ and $\mathbf{b}$ are:
\begin{align}
    \frac{\partial \mathcal{L}_s}{\partial W} &= \Delta_3(\mathbf{Q}_1\mathbf{V}^t)^\top, \quad
    \frac{\partial \mathcal{L}_s}{\partial \mathbf{b}} = \Delta_3\mathbf{1}_n.
\end{align}

\subsection{Derivation of Sampling Function Jacobian}

Given the trilinear interpolation sampling function:
\begin{equation}
    \mathrm{Sample}\left( \mathbf{H}_m^{t-1};\ \mathbf{R}^{t\rightarrow t-1} (\mathbf{P} + \mathbf{M}^{t}) \right),
\end{equation}
we aim to compute its Jacobian matrix $\mathbf{J}$ with respect to the sampling coordinates.
For a coordinate $\hat{p} = (x, y, z)^\top$ in $\mathbf{R}^{t\rightarrow t-1} (\mathbf{P} + \mathbf{M}^{t})$, the sampling function is defined as:
\begin{equation}
    \mathrm{Sample}\left( \mathbf{H}_m^{t-1};\ \hat{p} \right) = \mathbf{w}(\hat{p})^\top \mathbf{H}_m[\hat{p}],
\end{equation}
where \( \mathbf{w}(\hat{p}) \in \mathbb{R}^{8 \times 1} \) is the weight vector constructed using trilinear interpolation basis functions. $\mathbf{H}_m[\hat{\mathbf{p}}]$ represents the feature matrix at the eight corner points:
\begin{equation}
    \mathbf{H}_m[\hat{p}] = \begin{bmatrix}
        \mathbf{H}_m^{t-1}(i_0,\ j_0,\ k_0)^\top \\
        \mathbf{H}_m^{t-1}(i_0 + 1,\ j_0,\ k_0)^\top \\
        \mathbf{H}_m^{t-1}(i_0,\ j_0 + 1,\ k_0)^\top \\
        \mathbf{H}_m^{t-1}(i_0 + 1,\ j_0 + 1,\ k_0)^\top \\
        \mathbf{H}_m^{t-1}(i_0,\ j_0,\ k_0 + 1)^\top \\
        \mathbf{H}_m^{t-1}(i_0 + 1,\ j_0,\ k_0 + 1)^\top \\
        \mathbf{H}_m^{t-1}(i_0,\ j_0 + 1,\ k_0 + 1)^\top \\
        \mathbf{H}_m^{t-1}(i_0 + 1,\ j_0 + 1,\ k_0 + 1)^\top \\
    \end{bmatrix} ,
\end{equation}
 where \( i_0 = \lfloor x \rfloor \), \( j_0 = \lfloor y \rfloor \), and \( k_0 = \lfloor z \rfloor \). We define the fractional parts as \( d x = x - i_0 \), \( d y = y - j_0 \), and \( d z = z - k_0 \). The linear basis functions and their derivatives are:
\begin{align}
    \phi_0(e) &= 1 - e, & \phi_1(e) &= e, \\
    \phi_0'(e) &= -1, & \phi_1'(e) &= 1.
\end{align}
For each coordinate direction:
\begin{align}
    \mathbf{\phi}_x &= \begin{bmatrix} \phi_0(d x) \\ \phi_1(d x) \end{bmatrix}, & \mathbf{\phi}_x' &= \begin{bmatrix} \phi_0'(d x) \\ \phi_1'(d x) \end{bmatrix}, \\
    \mathbf{\phi}_y &= \begin{bmatrix} \phi_0(d y) \\ \phi_1(d y) \end{bmatrix}, & \mathbf{\phi}_y' &= \begin{bmatrix} \phi_0'(d y) \\ \phi_1'(d y) \end{bmatrix}, \\
    \mathbf{\phi}_z &= \begin{bmatrix} \phi_0(d z) \\ \phi_1(d z) \end{bmatrix}, & \mathbf{\phi}_z' &= \begin{bmatrix} \phi_0'(d z) \\ \phi_1'(d z) \end{bmatrix}.
\end{align}
The weight vector \( \mathbf{w}(\hat{p}) \) is then constructed using the Kronecker product \( \otimes \):
\begin{equation}
    \mathbf{w}(\hat{p}) = \mathbf{\phi}_x \otimes \mathbf{\phi}_y \otimes \mathbf{\phi}_z.
\end{equation}
The gradient of the weight vector is:
\begin{equation}
    \nabla_{\hat{p}} \mathbf{w}(\hat{p}) = \left[ \mathbf{\phi}_x' \otimes \mathbf{\phi}_y \otimes \mathbf{\phi}_z,\quad \mathbf{\phi}_x \otimes \mathbf{\phi}_y' \otimes \mathbf{\phi}_z\quad \mathbf{\phi}_x \otimes \mathbf{\phi}_y \otimes \mathbf{\phi}_z' \right],
\end{equation}
where each column corresponds to the partial derivative with respect to \( x \), \( y \), and \( z \), respectively.
The Jacobian matrix \( \mathbf{J} \) at position $\hat{p}$ of the sampling function with respect to \( \hat{p} \) is computed as:
\begin{equation}
    \mathbf{J}[\hat{p}] = \nabla_{\hat{p}} \mathrm{Sample}\left( \mathbf{H}_m^{t-1};\ \hat{p} \right) = \left( \nabla_{\hat{p}} \mathbf{w}(\hat{p}) \right)^\top \mathbf{H}_m[\hat{p}].
\end{equation}

\begin{table*}[!ht]

  \centering
  \resizebox{1.\textwidth}{!}{
  \begin{tabular}{l|c | c | c | c c c c c c c c c c c c c c c c c}
      \toprule
      Method
      
      & \rotatebox{90}{$\text{mIoU}$}
      & \rotatebox{90}{$\text{mIoU}_\text{D}$}
      & \rotatebox{90}{IoU}
      & \rotatebox{90}{\textcolor{otherscolor}{$\blacksquare$} others} 
      & \rotatebox{90}{\textcolor{barriercolor}{$\blacksquare$} barrier} %
      & \rotatebox{90}{\textcolor{bicyclecolor}{$\blacksquare$} bicycle} %
      & \rotatebox{90}{\textcolor{buscolor}{$\blacksquare$} bus} %
      & \rotatebox{90}{\textcolor{carcolor}{$\blacksquare$} car} %
      & \rotatebox{90}{\textcolor{constructcolor}{$\blacksquare$} cons. veh.} %
      & \rotatebox{90}{\textcolor{motorcolor}{$\blacksquare$} motor.} %
      & \rotatebox{90}{\textcolor{pedestriancolor}{$\blacksquare$} pedes.} %
      & \rotatebox{90}{\textcolor{trafficcolor}{$\blacksquare$} tfc. cone} %
      & \rotatebox{90}{\textcolor{trailercolor}{$\blacksquare$} trailer} %
      & \rotatebox{90}{\textcolor{truckcolor}{$\blacksquare$} truck} %
      & \rotatebox{90}{\textcolor{driveablecolor}{$\blacksquare$} drv. surf.} %
      & \rotatebox{90}{\textcolor{otherflatcolor}{$\blacksquare$} other flat} %
      & \rotatebox{90}{\textcolor{sidewalkcolor}{$\blacksquare$} sidewalk} %
      & \rotatebox{90}{\textcolor{terraincolor}{$\blacksquare$} terrain} %
      & \rotatebox{90}{\textcolor{manmadecolor}{$\blacksquare$} manmade} %
      & \rotatebox{90}{\textcolor{vegetationcolor}{$\blacksquare$} vegetation} \\ %
      \midrule
      BEVFormer~\cite{li2022bevformer} & 39.2  &37.2&- &5.0 & 44.9 & 26.2 & 59.7 & 55.1 & 27.9 & 29.1 & 34.3 & 29.6 & 29.1 & 50.5 & 44.4 & 22.4 & 21.5 & 19.5 & 39.3 & 31.1\\
      OSP~\cite{shi2024occupancy}   & 41.2 &37.0&-&11.0 & 49.0 & 27.7 & 50.2 & 56.0 & 23.0 & 31.0 & 30.9 & 30.3 & 35.6 & 41.2 & 82.1 & 42.6 & 51.9 & 55.1 & 44.8 & 38.2\\
       
        UniOCC~\cite{pan2023uniocc} & 39.7&-&- & - & - & - & - & - & - & - & - & - & - & - & - & - & - & - & - & - \\
        
        SurroundSDF~\cite{liu2024surroundsdf}   & 42.4&36.2&-&13.9&49.7&27.8&44.6&53.0&30.0&29.0&28.3&31.1&35.8&41.2&83.6&44.6&55.3&58.9&49.6&43.8 \\
        FlashOCC~\cite{yu2023flashocc}  & 32.0 &24.7&65.3 &6.2&39.6&11.3&36.3&44.0&16.3&14.7&16.9&15.8&28.6&30.9&78.2&37.5&47.4&51.4&36.8&31.4 \\
     
         COTR~\cite{ma2023cotr} &44.5 &38.6&75.0& 13.3 & 52.1 & 32.0 & 46.0 & 55.6 & 32.6 & 32.8 & 30.4 & 34.1 & 37.7 & 41.8 & 84.5 & 46.2 & 57.6 & 60.7 & 52.0 & 46.3 \\
   
ViewFormer~\cite{li2024viewformer}  & 41.9  &35.0 &70.2&12.9&50.1&28.0&44.6&52.9&22.4&29.6&28.0&29.3&35.2&39.4&84.7&49.4&57.4&59.7&47.4&40.6 \\
OPUS~\cite{wang2024opus}   &36.2 &33.3 &54.0 & 11.9 & 43.5 & 25.5 & 41.0 & 47.2 & 23.9 & 25.9 & 21.3 & 29.1 & 30.1 & 35.3 & 73.1 & 41.1 & 47.0 & 45.7 & 37.4 & 35.3\\

        \midrule
        BEVDetOcc-SF~\cite{huang2022bevdet4d, park2022time}  & 41.9 & 34.4 &75.1&12.1&50.0&22.1&43.9&53.9&\bf29.1&23.8&25.8&28.5&34.9&41.8&84.3&44.4&57.5&61.0&53.1&46.7 \\
         \rowcolor{pink!10}  \textbf{BEVDetOcc-GF}  & \textbf{43.6} & \textbf{36.0}&\textbf{77.8} &\bf12.6 & \bf51.5 & \bf24.0 & \bf46.2 & \bf55.8 & 26.8 & \bf26.3 & \bf27.3 & \bf30.8 & \bf37.6 & \bf43.4 & \bf84.7 &\bf 46.8 & \bf58.4 & \bf62.1 & \bf56.9 &\bf 50.7 \\
\midrule
FB-Occ~\cite{li2023fbocc}&39.8&34.2 &69.9&13.8&44.5&27.1&46.2&49.7&24.6&27.4&28.5&28.2&33.7&36.5&81.7&44.1&52.6&56.9&42.6&38.1 \\
       \rowcolor{pink!10}  \textbf{FB-Occ-GF }&\bf41.7&\bf35.8&\bf73.2 &\bf14.1 &\bf 47.6 &\bf 27.5 &\bf 46.8 & \bf52.0 & \bf26.8 &\bf 28.1 & \bf29.8 & \bf31.5 & \bf36.1 & \bf39.3 & \bf82.5 & \bf46.2 & \bf54.2 & \bf58.5 & \bf46.3 & \bf42.3 \\
       
       \midrule

ALOcc     &45.5 &39.3&75.3 &15.3&52.5&30.8&47.2&55.9&\bf32.7&33.3&32.4&36.2&\bf38.9&43.7&\bf84.9&48.5&\bf58.8&61.9&53.5&47.3 \\
              \rowcolor{pink!10}  \textbf{ALOcc-GF} &\bf46.5&\bf40.2&\bf77.4 &\bf 15.7 &\bf 53.1 &\bf 32.6 &\bf 48.5 &\bf 57.7 & 30.6 &\bf 34.1 &\bf 33.6 & \bf38.8 & 38.8 &\bf 45.2 & 84.8 &\bf 49.1 & 58.7 &\bf 62.4 &\bf 55.8&\bf49.9\\
       \midrule
       \midrule
        BEVDetOcc-SF   &47.1  & 41.2&76.8 
&15.6&\textbf{54.5}&33.0&53.4&58.1&32.6&34.1&32.3&35.4&40.5&46.6&85.8&50.1&60.4&63.3&55.9&49.7 \\
            \rowcolor{pink!10} \textbf{BEVDetOcc-GF}   &  \textbf{48.5} & \textbf{42.9} & \textbf{79.5} 
&15.6&54.2&\textbf{35.3}&\textbf{53.6}&\textbf{60.4}&\textbf{33.9}&\textbf{34.5}&\textbf{34.3}&\textbf{36.8}&\textbf{42.3}&\textbf{48.9}&\textbf{86.1}&50.1&\textbf{60.8}&\textbf{64.9}&\textbf{59.3}&\textbf{54.0} \\
            \midrule
            ALOcc   &50.6 &46.1 & 78.1 
&\textbf{17.0}&58.3&39.7&56.6&63.2&33.2&41.3&40.3&40.8&43.8&51.0&\textbf{87.0}&52.7&62.0&65.2&57.7&50.9  \\
            \rowcolor{pink!10} \textbf{ALOcc-GF}     & \textbf{51.8} & \textbf{47.4}& \textbf{79.9} 
&16.4&\textbf{58.4}&\textbf{40.1}&\textbf{58.6}&\textbf{64.6}&\textbf{34.0}&\textbf{43.2}&\textbf{41.9}&\textbf{43.5}&\textbf{44.6}&\textbf{52.6}&86.9&\textbf{53.2}&\textbf{62.3}&\textbf{65.9}&\textbf{60.4}&\textbf{54.0}\\
  \bottomrule
  \end{tabular}
  }
\caption{\textbf{Detailed per-class 3D semantic occupancy prediction results on Occ3D.} GDFusion consistently improves IoU for each class.}
\vspace{1ex}
  \label{tab:occ_cls}
\end{table*}

\begin{table*}[!ht]

  \centering
  \resizebox{1.\textwidth}{!}{
  \begin{tabular}{l|c | c | c | c c c c c c c c c c c c c c c c}
      \toprule
      Method
      
      & \rotatebox{90}{$\text{mIoU}$}
      & \rotatebox{90}{$\text{mIoU}_\text{D}$}
      & \rotatebox{90}{IoU}
      & \rotatebox{90}{\textcolor{barriercolor}{$\blacksquare$} barrier} %
      & \rotatebox{90}{\textcolor{bicyclecolor}{$\blacksquare$} bicycle} %
      & \rotatebox{90}{\textcolor{buscolor}{$\blacksquare$} bus} %
      & \rotatebox{90}{\textcolor{carcolor}{$\blacksquare$} car} %
      & \rotatebox{90}{\textcolor{constructcolor}{$\blacksquare$} cons. veh.} %
      & \rotatebox{90}{\textcolor{motorcolor}{$\blacksquare$} motor.} %
      & \rotatebox{90}{\textcolor{pedestriancolor}{$\blacksquare$} pedes.} %
      & \rotatebox{90}{\textcolor{trafficcolor}{$\blacksquare$} tfc. cone} %
      & \rotatebox{90}{\textcolor{trailercolor}{$\blacksquare$} trailer} %
      & \rotatebox{90}{\textcolor{truckcolor}{$\blacksquare$} truck} %
      & \rotatebox{90}{\textcolor{driveablecolor}{$\blacksquare$} drv. surf.} %
      & \rotatebox{90}{\textcolor{otherflatcolor}{$\blacksquare$} other flat} %
      & \rotatebox{90}{\textcolor{sidewalkcolor}{$\blacksquare$} sidewalk} %
      & \rotatebox{90}{\textcolor{terraincolor}{$\blacksquare$} terrain} %
      & \rotatebox{90}{\textcolor{manmadecolor}{$\blacksquare$} manmade} %
      & \rotatebox{90}{\textcolor{vegetationcolor}{$\blacksquare$} vegetation} \\ %
      \midrule
     
BEVFormer~\cite{li2022bevformer}  & 16.8 &14.2 & 30.5 &14.2&6.6&23.5&28.3&8.7&10.8&6.6&4.1&11.2&17.8&37.3&18.0&22.9&22.2&13.8&22.2\\

            TPVFormer~\cite{huang2023tri}   & 17.1 &14.0 & 30.9 &16.0&5.3&23.9&27.3&9.8&8.7&7.1&5.2&11.0&19.2&38.9&21.3&24.3&23.2&11.7&20.8

 \\
            SurroundOcc~\cite{wei2023surroundocc}  & 20.3 &18.4 & 31.5 &20.6&11.7&28.1&30.9&10.7&15.1&14.1&12.1&14.4&22.3&37.3&23.7&24.5&22.8&14.9&21.9  \\

            GaussianFormer~\cite{huang2024gaussianformer}   & 19.1 &17.3 & 29.8 &19.5&11.3&26.1&29.8&10.5&13.8&12.6&8.7&12.7&21.6&39.6&23.3&24.5&23.0&9.6&19.1  \\
            
            GaussianWorld~\cite{zuo2024gaussianworld}  & 22.1 &19.7 & 33.4 &21.4&14.1&27.7&31.8&13.7&17.4&13.7&11.5&15.1&23.9&43.0&24.9&28.8&26.7&15.7&24.7 \\
            
            \midrule
            BEVDetOcc~\cite{huang2022bevdet4d, park2022time}  & 17.5&14.1 & 29.2 
&18.1&2.1&25.5&29.5&11.6&9.5&7.0&4.4&7.3&20.1&40.4&21.3&26.3&23.8&11.5&21.6 \\
            \rowcolor{pink!10} \textbf{BEVDetOcc-GF}  & \textbf{20.8}&  \textbf{16.2} & \textbf{35.3} 
&\textbf{20.6}&\textbf{2.2}&\textbf{27.7}&\textbf{32.4}&\textbf{13.5}&\textbf{12.2}&\textbf{7.9}&\textbf{7.4}&\textbf{10.6}&\textbf{23.5}&\textbf{47.3}&\textbf{24.9}&\textbf{30.3}&\textbf{28.5}&\textbf{15.8}&\textbf{27.4}\\
            \midrule
            ALOcc-mini*~\cite{chen2024alocc}   & 21.5 &19.5 &31.5  &21.8&15.7&27.3&30.7&\textbf{12.7}&17.4&15.7&14.0&13.9&22.4&40.0&24.7&26.3&24.4&14.4&22.3\\
            \rowcolor{pink!10} \textbf{ALOcc-mini-GF}  &  \textbf{23.1} & \textbf{20.3} & \textbf{34.6}  &\textbf{22.2}&\textbf{16.0}&\textbf{27.9}&\textbf{32.7}&12.1&\textbf{18.9}&\textbf{16.6}&\textbf{15.3}&\textbf{14.5}&\textbf{23.9}&\textbf{46.0}&\textbf{28.2}&\textbf{29.0}&\textbf{26.6}&\textbf{15.8}&\textbf{23.7} \\
            \midrule
            ALOcc* \cite{chen2024alocc}  &24.0 &21.7 & 34.7 &23.8&17.4&28.0&32.9&17.0&\textbf{20.2}&17.2&16.9&15.4&25.5&41.8&26.7&28.3&27.0&18.6&27.0 \\
            \rowcolor{pink!10} \textbf{ALOcc-GF}  & \textbf{25.5} & \textbf{22.5}& \textbf{38.2}  
&\textbf{24.3}&\textbf{18.8}&\textbf{29.8}&\textbf{34.3}&\textbf{17.9}&19.6&\textbf{17.5}&\textbf{17.2}&\textbf{15.5}&\textbf{26.5}&\textbf{47.6}&\textbf{29.9}&\textbf{31.2}&\textbf{29.2}&\textbf{20.0}&\textbf{29.0}\\

  \bottomrule
  \end{tabular}
  }
\caption{\textbf{Detailed per-class 3D semantic occupancy prediction results on SurorundOcc.} }
\vspace{1ex}
  \label{tab:occ_cls_surroundocc}
\end{table*}

\begin{table*}[!ht]

  \centering
  \resizebox{1.\textwidth}{!}{
  \begin{tabular}{l|c|c | c | c | c c c c c c c c c c c c c c c c}
      \toprule
      Method
      & \rotatebox{90}{Input}
      
      & \rotatebox{90}{$\text{mIoU}$}
      & \rotatebox{90}{$\text{mIoU}_\text{D}$}
      & \rotatebox{90}{IoU}
      & \rotatebox{90}{\textcolor{barriercolor}{$\blacksquare$} barrier} %
      & \rotatebox{90}{\textcolor{bicyclecolor}{$\blacksquare$} bicycle} %
      & \rotatebox{90}{\textcolor{buscolor}{$\blacksquare$} bus} %
      & \rotatebox{90}{\textcolor{carcolor}{$\blacksquare$} car} %
      & \rotatebox{90}{\textcolor{constructcolor}{$\blacksquare$} cons. veh.} %
      & \rotatebox{90}{\textcolor{motorcolor}{$\blacksquare$} motor.} %
      & \rotatebox{90}{\textcolor{pedestriancolor}{$\blacksquare$} pedes.} %
      & \rotatebox{90}{\textcolor{trafficcolor}{$\blacksquare$} tfc. cone} %
      & \rotatebox{90}{\textcolor{trailercolor}{$\blacksquare$} trailer} %
      & \rotatebox{90}{\textcolor{truckcolor}{$\blacksquare$} truck} %
      & \rotatebox{90}{\textcolor{driveablecolor}{$\blacksquare$} drv. surf.} %
      & \rotatebox{90}{\textcolor{otherflatcolor}{$\blacksquare$} other flat} %
      & \rotatebox{90}{\textcolor{sidewalkcolor}{$\blacksquare$} sidewalk} %
      & \rotatebox{90}{\textcolor{terraincolor}{$\blacksquare$} terrain} %
      & \rotatebox{90}{\textcolor{manmadecolor}{$\blacksquare$} manmade} %
      & \rotatebox{90}{\textcolor{vegetationcolor}{$\blacksquare$} vegetation} \\ %
      \midrule

            TPVFormer \cite{huang2023tri} &C&7.8 & 11.7 &15.3 &9.7&4.5&11.5&10.7&5.5&4.6&6.3&5.4&6.9&6.9&14.1&9.8&8.9&9.0&9.9&8.5\\
             C-CONet \cite{wang2023openoccupancy} &C   &12.8 &10.6 & 20.1 &13.6&8.4&14.7&18.3&7.1&11.0&11.8&8.8&5.2&13.0&32.7&21.1&20.1&17.6&5.1&8.4
 \\
              C-OccGen \cite{wang2024occgen} &C &14.5 & 11.7 &23.4 &15.5&9.1&15.3&19.2&7.3&11.3&11.8&8.9&5.9&13.7&34.8&22.0&21.8&19.5&6.0&9.9
 \\

            \midrule
            ALOcc-2D* \cite{chen2024alocc} &C &15.7 &11.5 & 25.2 
&16.2&0.0&14.2&20.4&10.0&12.3&12.5&11.2&7.7&14.9&35.2&23.8&23.4&21.1&11.8&16.0 \\
            \rowcolor{pink!10} \textbf{ALOcc-2D-GF} &C   & \textbf{17.9} & \textbf{13.7}& \textbf{28.6}  
&\textbf{17.4}&\textbf{11.5}&\textbf{15.2}&\textbf{21.7}&\textbf{10.3}&\textbf{13.8}&\textbf{12.9}&\textbf{11.4}&\textbf{8.1}&\textbf{15.7}&\textbf{41.3}&\textbf{27.2}&\textbf{26.6}&\textbf{23.7}&\textbf{12.6}&\textbf{17.4}\\
 \midrule
            \midrule
            L-CONet \cite{wang2023openoccupancy} &L   &15.8 &10.8 & 30.9 &18.0&3.9&14.2&18.7&8.3&6.3&11&5.8&14.1&14.3&35.3&20.2&21.5&20.9&19.2&23.0 \\
            L-OccGen \cite{wang2024occgen} &L   &16.8 &11.8 & 31.6 &18.8&5.1&14.8&19.6&7.0&7.7&11.5&6.7&13.9&14.6&36.4&22.1&22.8&22.3&20.6&24.5 \\
            3DSketch$^{\dagger}$ \cite{chen20203d}  &C\&D  &10.7 &7.4 & 25.6 &12.3&5.2&10.3&12.1&7.1&4.9&5.5&6.9&8.4&7.4&21.9&15.4&13.6&12.1&12.1&21.2  \\
              AICNet$^{\dagger}$ \cite{li2020anisotropic}  &C\&D  &10.6 &7.4 & 23.8 &11.8&4.5&12.1&12.7&6.0&3.9&6.4&6.3&8.4&7.8&24.2&13.4&13.0&11.9&11.5&20.5 \\
            M-CONet \cite{wang2023openoccupancy} &C\&L  &20.1 &18 & 29.5 &23.3&16.1&22.2&24.6&13.3&20.1&21.2&14.4&17.0&21.3&31.8&22.0&21.8&20.5&17.7&20.4
 \\
             M-OccGen \cite{wang2024occgen} &C\&L  &22.0 &20 &30.3 &24.9&16.4&22.5&26.1&14.0&20.1&21.6&14.6&17.4&21.9&35.8&24.5&24.7&24.0&20.5&23.5 \\
              Co-Occ \cite{pan2024co} &C\&L &21.9 &19.5 &30.6 &26.5&16.8&22.3&27.0&10.1&20.9&20.7&14.5&16.4&21.6&36.9&23.5&25.5&23.7&20.5&23.5 \\
             OccLoff \cite{zhang2025occloff} &C\&L  &22.9 &21.1 &31.4 &26.7&17.2&22.6&26.9&16.4&22.6&24.7&16.4&16.3&22.0&37.5&22.3&25.3&23.9&21.4&24.2 \\
             \midrule
             
            ALOcc-2D* \cite{chen2024alocc}  &C\&D   &22.4 &20.3 & 30.4 
&24.1&16.9&20.7&26.7&14.2&22.2&25.7&16.1&14.7&21.7&35.2&24.7&24.7&23.4&22.4&24.9  \\
            
            \rowcolor{pink!10} \textbf{ALOcc-2D-GF}   &C\&D & \textbf{24.5} & \textbf{21.6}& \textbf{34.5}  
&\textbf{25.7}&\textbf{17.1}&\textbf{21.6}&\textbf{28.3}&\textbf{15.0}&\textbf{24.0}&\textbf{26.9}&\textbf{16.6}&\textbf{17.1}&\textbf{23.0}&\textbf{42.0}&\textbf{29.0}&\textbf{28.4}&\textbf{26.7}&\textbf{23.9}&\textbf{26.8}\\

  \bottomrule
  \end{tabular}
  }
\caption{\textbf{Detailed per-class 3D semantic occupancy prediction results on OpenOccupancy.} }
\vspace{1ex}
  \label{tab:occ_cls_openoccupancy}
\end{table*}

\section{Additional Resutls}

\subsection{Detailed Per-Class Semantic Occupancy Prediction}
As shown in ~\cref{tab:occ_cls,tab:occ_cls_surroundocc,tab:occ_cls_openoccupancy}, we present the IoU for all categories. GDFusion consistently improves the performance of the baselines across most categories, demonstrating the generality of our approach. Notably, our method achieves significant improvements in background categories such as \textit{driveable surface}, \textit{vegetation} and \textit{manmade}, while also providing considerable gains for dynamic object categories like \textit{car} and \textit{pedestrian}.

\begin{table*}[t]

    \setlength{\tabcolsep}{5pt}
    \centering
    \setlength{\tabcolsep}{2.8mm}{
    \scalebox{0.80}{
    \begin{tabular}{l|c|c|c|ccc}
        \toprule
        Method & Backbone & Input Size & {RayIoU} & \multicolumn{3}{c}{RayIoU\textsubscript{1m, 2m, 4m}} \\
        \midrule
        \midrule

        RenderOcc \cite{pan2024renderocc} & Swin-Base & 512$\times$1408  & 19.5 & 13.4 & 19.6 & 25.5 \\
        SparseOcc \cite{liu2023fully}          & ResNet-50    & 256$\times$704   & {36.1} & 30.2 & 36.8 &41.2  \\
        Panoptic-FlashOcc \cite{yu2024panoptic}          & ResNet-50    & 256$\times$704   & 38.5 &32.8 &39.3 &43.4  \\
        OPUS \cite{wang2024opus}          & ResNet-50    & 256$\times$704   & 41.2 & 34.7 &42.1 &46.7  \\
        \midrule
        \midrule
        BEVDetOcc-SOLO \cite{huang2022bevdet4d, park2022time}    & ResNet-50    & 256$\times$704 & 35.2 & 31.2 & 35.9 & 38.4 \\
        \rowcolor{pink!10} \textbf{BEVDetOcc-GF}   & ResNet-50    & 256$\times$704  & \textbf{36.6}{\color{red2}~$\uparrow$1.4}   &  \textbf{32.6}{\color{red2}~$\uparrow$1.4}   &   \textbf{37.3}{\color{red2}~$\uparrow$1.4}  & \textbf{39.9}{\color{red2}~$\uparrow$1.5} \\
        \midrule
        FB-Occ \cite{li2023fbocc}      & ResNet-50    & 256$\times$704 & 39.0 & 33.0 & 40.0 & 44.0 \\
        \rowcolor{pink!10} \textbf{FB-Occ-GF}     & ResNet-50    & 256$\times$704  & \textbf{40.6}{\color{red2}~$\uparrow$1.6}   &  \textbf{35.0}{\color{red2}~$\uparrow$2.0}   &  \textbf{41.5}{\color{red2}~$\uparrow$1.5}   &  \textbf{45.3}{\color{red2}~$\uparrow$1.3} \\
        \midrule
        ALOcc \cite{chen2024alocc} & ResNet-50 & 256$\times$704 & 43.7 & 37.8 & 44.7 & 48.8 \\
        \rowcolor{pink!10} \textbf{ALOcc-GF}   & ResNet-50    & 256$\times$704  &  \textbf{44.1}{\color{red2}~$\uparrow$0.4}   &  \textbf{38.2}{\color{red2}~$\uparrow$0.4}   &  \textbf{45.0}{\color{red2}~$\uparrow$0.3}   &  \textbf{49.2}{\color{red2}~$\uparrow$0.4} \\
        \bottomrule
    \end{tabular}
    }
    }
     \caption{\textbf{Evaluation of 3D semantic occupancy prediction on the Occ3D benchmark without using the camera-visible mask, assessed using RayIoU metrics.} Relative improvements are highlighted with red arrows {\color{red2}~$\uparrow$}. The integration of GDFusion demonstrates consistent and substantial performance enhancements across the baseline methods.}
     \label{tab:sota-womask}
\end{table*}

\subsection{Performance Evaluated with RayIoU}
Recently, \citet{liu2023fully} proposed using RayIoU to evaluate semantic occupancy prediction, providing an alternative perspective to the evaluation system in Occ3D \cite{tian2024occ3d}. The experiments in \cref{tab:sota-womask} showcase the significant impact of GDFusion on advancing 3D semantic occupancy prediction under training conditions without a camera-visible mask.
The results in \cref{tab:sota-womask} clearly demonstrate the effectiveness of integrating GDFusion, which consistently improves the performance of baseline models across RayIoU metrics, as indicated by the red arrows marking relative improvements. These findings underscore the effectiveness of GDFusion in leveraging valuable information embedded within temporal cues, thereby enhancing both the geometric coherence and semantic precision of reconstructed scenes. As a result, GDFusion enables more reliable and accurate occupancy predictions.

\begin{table}[t]
        
        \centering
        \small
        \setlength{\tabcolsep}{4pt}
        \resizebox{0.52\linewidth}{!}{
        \begin{tabular}{l|ccc}
            \toprule
            Method  & mIoU &  mIoU\textsubscript{D} &  IoU  \\
            \midrule
            \midrule
            Baseline &38.0 & 31.0 & 71.1 \\
            Our Vox his &41.8 & 34.0 & 76.5\\
         RWKV  & 38.2 & 31.3& 72.2  \\
         RWKV + Our Scene, Motion, Geometry His &41.9 & 35.0& 76.1  \\
         xLSTM  & 39.9 & 32.7 & 74.4 \\
         xLSTM + Our Scene, Motion, Geometry His & 40.9 &33.6   &76.1   \\
         Mamba &41.2 &33.9   &75.0   \\
         Mamba + Our Scene, Motion, Geometry His &42.4 &34.6   &76.8   \\

         \midrule
         RWKV + Our Vox his & 41.7 &33.5  &76.5   \\
         RWKV + Our All His & 43.3 & 35.8& 77.8  \\
         xLSTM + Our Vox his & 42.2 & 34.7  & 76.4  \\
         xLSTM + Our All his & 43.3 & 35.7 &77.5 \\
         Mamba + Our Vox his &41.9 & 34.4 & 76.4 \\
         Mamba + Our All his &43.1 &  35.2 &  77.8 \\
         Our Full &43.3 &35.3   &77.8  \\
            \bottomrule
        \end{tabular}
        }
        \caption{\textbf{Extension study on integrating modern RNN methods with our method.}}
        \label{tab:modernrnn}
    \end{table}

\subsection{Results on Modern RNNs}

Our voxel-level history fusion module can be directly replaced with modern RNN methods such as RWKV \cite{peng2023rwkv}, xLSTM \cite{beck2024xlstm}, and Mamba \cite{gu2023mamba}. In ~\cref{tab:modernrnn}, we evaluate the performance of combining these modern RNN methods with our approach on Occ3D. For RWKV, xLSTM, and Mamba, we used a single-layer model for each respective structure. From the table, we draw several conclusions: 
First, a standalone modern RNN method cannot outperform our voxel-level history fusion module. Second, combining our proposed auxiliary temporal modules with a modern RNN method yields significant improvements. Third, while integrating modern RNN methods with our voxel-level fusion approach and other temporal fusion modules does result in improvements, it does not outperform the setting without modern RNNs (\ie, Our Full). We hypothesize that modern RNNs, which are designed for long-sequence context understanding in tasks like natural language processing, do not show clear advantages in the relatively short sequences of the nuScenes dataset when used solely for temporal fusion. Utilizing modern RNNs for integrating both spatial and temporal dimensions could be explored as a direction for future research.

\begin{table}[t]
        
        \centering
        \small
        \setlength{\tabcolsep}{4pt}
        \resizebox{0.63\linewidth}{!}{
        \begin{tabular}{l|ccc}
            \toprule
            Position  & mIoU &  mIoU\textsubscript{D} &  IoU  \\
            \midrule
            \midrule
            Before Depth Net  &42.1&34.6&76.6\\
            Before Voxel-level His Fusion&42.4&34.8&76.8 \\
         After Voxel-level His Fusion&42.5&34.8&77.0   \\
         After Volume Encoder  &42.1&34.5&76.6\\
         After Voxel-level His Fusion + Before Voxel-level His Fusion &42.4&34.6&77.2   \\
         After Voxel-level His Fusion + After Volume Encoder  &42.4&34.7&77.1 \\
         All &42.0 &34.2  &76.9   \\

            \bottomrule
        \end{tabular}
        }
        \caption{\textbf{Ablation study \textit{\wrt} the position of scene-level history fusion in the framework.}}
        \label{tab:scene}
    \end{table}

\subsection{Experiments on the Position of Scene-Level History Fusion in the Framework}

In \cref{tab:scene}, we investigate the impact of the position of scene-level history fusion on network performance on Occ3D. The baseline model employs voxel-level history fusion. Specifically, we consider several positions within the vision-based semantic occupancy network architecture: before the depth network, before voxel-level history fusion, after voxel-level history fusion, and after the volume encoder, as well as multiple positions for scene-level fusion. The results in the table indicate that scene-level fusion before the depth network or after the volume encoder performs worse compared to fusion before or after voxel-level history fusion. The poor performance of fusion before the depth network is attributed to the fusion occurring in the 2D modality, where the difference in data structure limits its effectiveness compared to direct fusion in the 3D modality. The inferior performance of fusion after the volume encoder is primarily because scene-level history fusion acts similarly to domain adaptation, which is beneficial for generating domain-independent features, favoring subsequent network encoding. Therefore, fusion before the volume encoder, which is a densely encoding module, proves to be more advantageous. We also experimented with multiple positions for scene-level fusion, but the results showed no significant advantage over a single fusion position. Consequently, we only perform fusion after the voxel-level history fusion module in the final method.

\begin{table*}[t]
  \centering
  \small
  \setlength{\tabcolsep}{4pt}
 
      \resizebox{0.23\linewidth}{!}{
        \begin{tabular}{l|ccc}
          \toprule
          $\eta_m$ & 0.001 & 0.01 & 0.1 \\
          \midrule
          mIoU & 42.5 & 42.5 & 42.4 \\
          mIoU\textsubscript{D} & 32.4 & 32.4 & 32.4 \\
          IoU & 76.6 & 76.5 & 76.4 \\
          \bottomrule
        \end{tabular}
      }
      \caption{\textbf{Parameter study on $\eta_m$.}}
      \label{tab:motion}

\end{table*}

\begin{table}[t]
  \centering
  \small
  \setlength{\tabcolsep}{4pt}
  \resizebox{0.39\linewidth}{!}{
    \begin{tabular}{l|c|ccc}
      \toprule
      Method & $\eta_s$ & mIoU & mIoU\textsubscript{D} & IoU \\
      \toprule
      \multirow{4}{*}{Scene His} & 0.1 & 42.5 & 34.8 & 77.0 \\
                                 & 1.0 & 42.4 & 34.2 & 77.3 \\
                                 & 10  & 42.3 & 34.1 & 77.5 \\
                                 & 100 & 42.1 & 34.3 & 77.2 \\
      \midrule
      \multirow{4}{*}{w/o Gradient on $\gamma, \beta$} & 0.1 & 42.3 & 34.4 & 77.1 \\
                                                        & 1.0 & 42.4 & 34.5 & 76.9 \\
                                                        & 10  & 42.5 & 34.9 & 77.0 \\
                                                        & 100 & 42.1 & 34.3 & 76.6 \\
      \midrule
      \multirow{2}{*}{Linear $\rightarrow$ MLP} & 0.01 & 42.5 & 35.1 & 76.9 \\
                                               & 0.1  & 42.4 & 34.9 & 77.0 \\
      \bottomrule
    \end{tabular}
  }
  \caption{\textbf{Ablation study \textit{w.r.t.} the structure of the scene-level history fusion module.} The first row indicates the selected structure for scene-level history fusion. The second row shows the case where only linear layer parameters $\mathbf{W}$ and $\mathbf{b}$ are updated during scene-level history fusion. The third row represents extending the linear layer to an MLP by adding additional parameters.}
  \label{tab:scene_ab}
\end{table}

\subsection{Module Structure and Impact of Parameters $\eta_s$ and $\eta_m$}
In \cref{tab:scene_ab}, we explore different structural choices for scene-level history fusion on Occ3D. The baseline model utilizes voxel-level temporal fusion. Jointly applying history fusion to both linear and LayerNorm parameters improves IoU to some extent, likely due to the influence of LN parameters on background categories, which occupy a large proportion of the scene. In the third row, we replace the linear layer with an MLP, which means using more parameters to store historical scene information. However, experimental results indicate no significant gains, so we ultimately use only the linear and LN layers to store scene information.

In \cref{tab:scene_ab} and \cref{tab:motion}, we evaluate the impact of the learning rate parameters $\eta_s$ and $\eta_m$ in scene-level temporal fusion and temporal motion fusion on model performance. The results demonstrate that our method is relatively insensitive to these hyperparameters.

\begin{figure*}[t]
    \centering
    \setlength{\abovecaptionskip}{0pt}
    \includegraphics[width=1\linewidth]{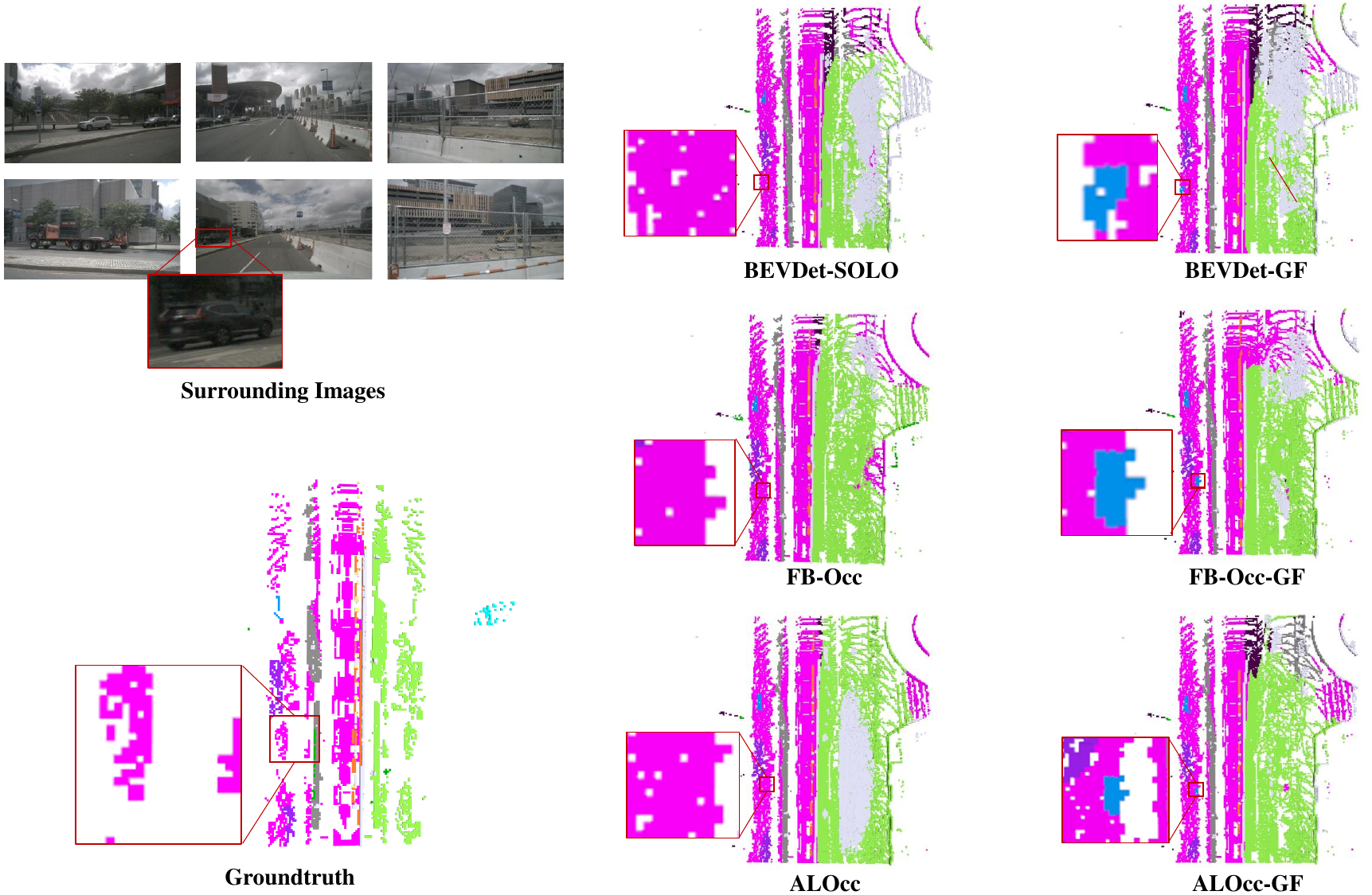}
    \caption{Qualitative comparison between BEVDetOcc-SF, FBOcc, and ALOcc, each enhanced by our GDFusion method on Occ3D. The top row in the leftmost column shows the input images, presented in the following order: camera front left, camera front, camera front right, camera back left, camera back, and camera back right. The bottom row in the leftmost column displays the ground-truth semantic occupancy. The middle section illustrates the results of the three baselines, while the rightmost column presents the results after incorporating our method. Key areas are highlighted with red boxes for emphasis.}
    \vspace{1ex}
    \label{fig:occ}
\end{figure*}
\subsection{Qualitative Analysis}
Figure~\ref{fig:occ} presents the qualitative results of our method. Notably, none of the three baselines were able to predict the presence of the car in the image, and even the ground truth lacks annotations for this car due to the sparse nature of multi-frame LiDAR-aggregated groundtruth. However, after incorporating our method, all three approaches successfully detected the car, demonstrating the robustness and generalizability of our approach, even in scenarios where the groundtruth is incomplete.

\end{document}